\documentclass{article}

\newif\ifarXiv
\arXivtrue

\PassOptionsToPackage{sort, compress, numbers}{natbib}

 \usepackage{iclr2027_conference}
 \usepackage{times}

\ifarXiv
    \usepackage{etoolbox}
    \usepackage[noblocks]{authblk}

    \makeatletter
    \def\maketitle{{\renewenvironment{tabular}[2][]{\begin{flushleft}}{\end{flushleft}}\AB@maketitle}}
    \makeatother
    \iclrfinalcopy
\fi

\usepackage[utf8]{inputenc}
\usepackage[T1]{fontenc}
\usepackage{graphicx}
\usepackage{caption}
\usepackage{subcaption}
\usepackage{hyperref}
\usepackage{url}
\usepackage{booktabs}
\usepackage{tabularx}
\usepackage{multirow}
\usepackage{wrapfig}
\usepackage{amsmath}
\usepackage{amsfonts}
\usepackage{amssymb}
\usepackage{nicefrac}
\usepackage{microtype}
\usepackage{silence}
\usepackage{xcolor}
\usepackage{paralist}
\usepackage[capitalise, nameinlink]{cleveref}

\usepackage{enumitem}
\usepackage{titletoc}
\usepackage[hang, flushmargin]{footmisc}

\usepackage{array} 
\usepackage{mathtools}

\usepackage{babel}

\usepackage{tcolorbox}
\usepackage{float}

\definecolor{OIOrange}    {RGB}{230, 159,   0}
\definecolor{OISkyBlue}   {RGB}{ 86, 180, 233}
\definecolor{OIGreen}     {RGB}{  0, 158, 115}
\definecolor{OIYellow}    {RGB}{240, 228,  66}
\definecolor{OIBlue}      {RGB}{  0, 114, 178}
\definecolor{OIVermillion}{RGB}{213,  94,   0}
\definecolor{OIRose}      {RGB}{204, 121, 167}

\tcbset{
  boxsep     = 1.0mm,
  colframe   = OIBlue,
  colback    = OIBlue!10,
  left       = 2.0mm,
  right      = 2.0mm,
sharp corners,
}

\hypersetup{
    colorlinks = true,
    urlcolor   = OIBlue,
    citecolor  = OIBlue,
    linkcolor  = OIBlue,
}

\title{
Measuring Structure in Graph Benchmark Datasets Using Graph Invariants
}

\ifarXiv
    \author[1]{Richard von Moos}
    \author[2]{Mathieu Alain}
    \author[1]{Bastian Rieck}
    
    \affil[1]{AIDOS Lab, University of Fribourg}
    \affil[2]{University College London} 
\else
    \author{}
\fi

\begin{document}

\maketitle

\ifarXiv
    \lhead{Preprint. Under review.}
\fi

\begin{abstract}Progress in graph learning is hindered by benchmark practices that conflate the contributions of node features and graph structure, making it hard to tell whether a model actually learns from the graph, or whether it even needs to.
We propose measuring this using \emph{graph invariants}, i.e., permutation-invariant, task-agnostic structural descriptors.
Our analysis on $26$ datasets substantiates three tacit assumptions in graph learning, namely that
\begin{inparaenum}[(i)]
       \item a curated subset of invariants is more expressive than standard GNNs,
       \item benchmark datasets exhibit structural heterogeneity even when originating from the same domain, and
       \item simple models are often competitive with, and sometimes exceed, approaches based on transformers or message passing.
   \end{inparaenum}
We thus posit that graph invariants should become a standard tool for measuring graph learning task complexity and the relevance of graph structure.
\end{abstract}

\section{Introduction}

Graphs offer a parsimonious representation for a wide variety of datasets and tasks~\citep{Velickovic23a}, giving rise to efficient machine learning models that are predominantly based on the message-passing paradigm~\citep{Corso24a}.
However, whereas a plethora of graph learning models exists, the \emph{datasets} used to compare and evaluate them have received far less attention, leading to stagnating or even decreasing performance~\citep{Bechler-Speicher25a}.
Recent work thus started pointing out issues concerning datasets and
    evaluation practices~\citep{Bechler-Speicher25a, Coupette25a, Palowitch22a,
    Ballester25a, toenshoff2024where}, including
\begin{inparaenum}[(i)]
        \item small dataset sizes, which often prevent making statistically significant statements concerning predictive performance,
        \item a lack of structural diversity within datasets, thus preventing model generalization, and
        \item a disproportionate focus on metrics like \emph{expressivity} that are not a priori aligned \emph{with} or even required \emph{for} a specific task.
    \end{inparaenum}

    In the absence of an agreed-upon methodology for assessing these issues, our
    work emphasizes assessing the graph structure and its impact on tasks.
Since common graph learning models tend to intertwine graph structure with node and edge features, we adopt a principled approach that uses \emph{graph
    invariants} to create structural fingerprints of a graph.
Within the field of \emph{complex network analysis}~\citep{Newman03a}, specific types of invariants have already been successfully used. This includes, for instance, motif counts for structural characterizations~\citep{Yin26a} or topological indices for quantifying network complexity~\citep{Dehmer15a, Dehmer17a}.
In graph learning, by contrast, invariants are less prevalent and their use is typically restricted to serving as an additional inductive bias of a model, tailored to improve predictive performance; examples include \emph{homomorphism counts}~\citep{jin24a, Bao25a}, \emph{clique counts}~\citep{Bouritsas23a} and \emph{Laplace spectra}~\citep{alain26}.
Our work substantially expands on this perspective by using a comprehensive and diverse set of graph invariants that can serve as computable proxies for graph structure.
This enables us to study to what extent structure is relevant for solving graph-level tasks in current graph benchmark datasets.
Next to raising questions about dataset quality, we contribute a novel experimental evaluation methodology based on \emph{graph invariant baselines}.
We show that such baselines are often competitive, providing a simple diagnostic tool to discover model--task misalignment.

\section{Using graph invariants in graph learning}
\label{sec:preliminaries}
    \paragraph{Invariants.}
    \label{sec:invariants}
        Although local aggregation is central to architectures such as graph convolutional networks \citep{Kipf2016} and graph attention networks \citep{velickovic2018graph}, incorporating global structural information remains challenging. One natural way to encode such information is through graph invariants, i.e., interpretable, permutation-invariant quantities.
Formally, let $\mathcal{G}$ denote the space of finite undirected graphs without self-loops. A function $I\colon\mathcal{G}\to\mathbb{R}$ is a graph invariant if, for any two isomorphic graphs $G,H\in\mathcal{G}$, we have $I(G)=I(H)$.
For example, the \emph{Wiener index} \citep{Wiener1947} measures the spread of a connected graph $G=(V,E)$ by summing shortest-path distances over all pairs of vertices. Path graphs have a large Wiener index, whereas complete graphs have a small Wiener index. The Wiener index has therefore been widely used as a simple descriptor of molecular structure and biological activity~\citep{Rouvray2022}. Numerous other such graph invariants~(see \Cref{app:invariants} for an overview of the ones used in this paper) provide a natural lens through which to study graph expressivity. While a \emph{single} invariant is usually insufficient to distinguish all non-isomorphic graphs, families of invariants from different domains can capture increasingly rich structural information. Thus, graph invariants offer a concrete way to encode global graph structure rather than only local aggregation patterns. 
        Similarly, local topological profiles have previously served as structural descriptors~\citep{Adamczyk23a}.
        
    \paragraph{Using invariants with machine learning models.}
To use graph invariants as structural fingerprints, we require an
        approach that lets us measure their expressive and predictive
        power on a multitude of datasets in a reproducible fashion while
        remaining resource-efficient. At the same time, a suitable
        model should perform at comparable levels \emph{across} datasets
        to make for a suitable baseline.  An obstruction to all three, i.e., 
        reproducibility, comparability, and ease of application, is that
        model performance usually depends on the choice of hyperparameters
        that themselves depend on the specific task. Hence, even if finding a
        suitable set of hyperparameters is possible, it may take an
        infeasible amount of resources.
To reduce the number of manual choices as much as possible while
        remaining flexible with respect to the choice of dataset, we
        thus propose using XGBoost~\citep{chen16a} as a base model,
        providing feature vectors built from invariants as inputs. 
Compared to GNNs, XGBoost is computationally more efficient,
        exposes fewer hyperparameters, and is therefore considerably
        easier to tune.
Specifically, XGBoost permits fully-automated hyperparameter
        tuning via Optuna~\citep{akiba19a}, resulting in high flexibility
        and reproducibility; in fact, we can easily apply the same
        tuning procedure to new datasets, obtaining an overall
        \emph{robust} and flexible baseline model.
By contrast, the higher complexity of GNNs results in a substantially slower tuning procedure; for our experiments in \Cref{sec:relevance_of_structure}, we observe that GNNs finish at most a single tuning iteration in the time it takes to finish tuning XGBoost.

    \paragraph{Selecting invariants.}
        Another obstruction to a reproducible invariant baseline model
        is the selection of invariants for a specific task. We ideally
        would want a \emph{small}, maximally \emph{independent} set of
        invariants, each of which is \emph{computationally efficient}. At the
        same time, we want invariants to constitute a comprehensive structural summary
        of a graph.
We first prioritize comprehensiveness and therefore include even
        strongly-correlated invariants~(cf.\
        \Cref{fig:mean_min_corr_heatmap}) in our experiments, since we observe that this
        does not markedly reduce predictive performance in \emph{any} of
        the tasks we investigate.
This approach, while computationally inefficient, constitutes a
        practical way of gauging upper bounds for predictive performance
        while enabling us to avoid manually curating invariants per
        task/dataset~(which would contradict the idea of a
        \emph{general} baseline model based on graph invariants).
Practically, we observe that invariant calculation~(which has to
        be performed only \emph{once} per dataset) is feasible for
        typical graph-level tasks~(\Cref{tab:invariant_comp_time}) and
        can in fact be trivially parallelized.
However, to accommodate datasets in which the size of individual
        graphs makes the calculation of certain invariants infeasible,
        we also define smaller subsets of invariants and empirically
        show that they provide similar insights~(cf.\ \Cref{app:regimes_and_invariant_sets}).

\section{Experiments: How invariants support graph learning tasks}
\label{sec:Experiments}

In the following, we assess the utility of invariants in three different scenarios.
The commonality of all subsequent experiments is that, whenever possible, we precompute invariants for \emph{all} graphs of a dataset.
Subsequently, we use them as static features for expressivity analysis as well as augmentations of graph-level representations for graph-level prediction tasks.

Our main questions are:
\begin{enumerate}[
    label=Q\arabic*:,
    leftmargin=3em,
    labelsep=0.5em,
    itemsep=0pt,
    topsep=2pt]
  \item How \emph{expressive} are invariants?
  \item To what extent do invariants \emph{characterize} existing datasets?
  \item Do invariants \emph{capture} the relevance of graph structure?
\end{enumerate}

\subsection{Invariants are expressive}
\label{sec:expressivity}

    Certain types of invariants~(e.g., \emph{local homomorphism counts}, \emph{Laplacian eigenvalues}, \emph{molecular indices}) have already been shown to capture relevant structural information for graph learning tasks~\citep{alain24, jin24a, dwivedi20a, moriwaki18a}.
We thus hypothesize that the combination of multiple invariants should be capable of capturing structural information across many classes of graphs.
To quantify this, we investigate the ability of invariants to differentiate between non-isomorphic pairs of graphs.

    Our experiment uses the BREC dataset~\citep{wang24a}, which consists of $400$ pairs of graphs that are not distinguishable by \mbox{$1$-WL}, the Weisfeiler--Lehman test for graph isomorphism~\citep{morris23a}.
This dataset is commonly used to assess the realized expressivity of graph learning models.
To determine whether an encoding method can distinguish between two graphs, the \emph{encodings} of both graphs are compared to \emph{encodings} of their isomorphic relabellings, with the requirement that differences between the pair have to be consistently large, whereas differences between their relabelled versions have to be consistently small~\citep{wang24a}.
Since graph invariants are \emph{invariant} under any graph isomorphism, this criterion is substantially simplified and only requires us to check whether two invariants have the \emph{same} value for a pair of non-isomorphic graphs.
    \begin{table*}[btp]
\centering
\caption{Excerpt from \citet{wang24a} with six additional rows showing the performance of invariant-based approaches for the full invariant set $\texttt{I}$, the expressive subset $\texttt{S}$, and their efficient-regime counterparts $\texttt{I}_{\mathrm{eff}}$ and $\texttt{S}_{\mathrm{eff}}$~(cf.\ \Cref{tab:invariants_subsets}). ``(raw)'' denotes differentiated pairs based on differences with an error tolerance of $10^{-6}$ between invariant vectors. ``(MLP)'' denotes an MLP trained on invariant vectors, following \citet{wang24a}.
For this expressivity experiment, we observe \emph{equal performance} between both types of approaches.
}
\label{tab:expressivity_results}
\small
\setlength{\tabcolsep}{4pt}
\resizebox{\textwidth}{!}{\begin{tabular}{llcccccccccc}
\toprule
 & & \multicolumn{2}{c}{Basic Graphs (60)} & \multicolumn{2}{c}{Regular Graphs (140)} & \multicolumn{2}{c}{Extension Graphs (100)} & \multicolumn{2}{c}{CFI Graphs (100)} & \multicolumn{2}{c}{Total (400)} \\
\cmidrule(lr){3-4} \cmidrule(lr){5-6} \cmidrule(lr){7-8} \cmidrule(lr){9-10} \cmidrule(lr){11-12}
Type & Model & Number & Accuracy & Number & Accuracy & Number & Accuracy & Number & Accuracy & Number & Accuracy \\
\midrule
\multirow{3}{*}{Non-GNNs}
 & 3-WL    & 60 & 100\%  & 50  & 35.7\% & 100 & 100\% & 60 & 60.0\% & 270 & 67.5\% \\
 & SPD-WL  & 16 & 26.7\% & 14  & 11.7\% & 41  & 41\%  & 12 & 12\%   & 83  & 20.8\% \\
 & $N_2$   & 60 & 100\%  & 138 & 98.6\% & 100 & 100\% & 0  & 0\%    & 298 & 74.5\% \\
\midrule
\multirow{5}{*}{Subgraph GNNs}
 & NGNN          & 59 & 98.3\% & 48  & 34.3\% & 59  & 59\%  & 0  & 0\%  & 166 & 41.5\% \\
 & DSS-GNN       & 58 & 96.7\% & 48  & 34.3\% & 100 & 100\% & 15 & 15\% & 221 & 55.2\% \\
 & SUN           & 60 & 100\%  & 50  & 35.7\% & 100 & 100\% & 13 & 13\% & 223 & 55.8\% \\
 & GNN-AK        & 60 & 100\%  & 50  & 35.7\% & 97  & 97\%  & 15 & 15\% & 222 & 55.5\% \\
 & $I^2$-GNN     & 60 & 100\%  & 100 & 71.4\% & 100 & 100\% & 21 & 21\% & 281 & 70.2\% \\
\midrule
\multirow{1}{*}{$k$-WL GNNs}
 & PPGN              & 60 & 100\% & 50 & 35.7\% & 100 & 100\% & 23 & 23\% & 233 & 58.2\% \\
\midrule
Substructure GNNs    & GSN        & 60 & 100\%  & 99 & 70.7\% & 95 & 95\% & 0  & 0\%  & 254 & 63.5\% \\
\midrule
\multirow{2}{*}{Random GNNs}
 & DropGNN     & 52 & 86.7\% & 41 & 29.3\% & 82 & 82\% & 2 & 2\% & 177 & 44.2\% \\
 & OSAN        & 56 & 93.3\% & 8  & 5.7\%  & 79 & 79\% & 5 & 5\% & 148 & 37\%   \\
\midrule
Transformer GNNs     & Graphormer & 16 & 26.7\% & 12 & 8.6\%  & 41 & 41\% & 10 & 10\% & 79 & 19.8\% \\
\midrule
\multirow{5}{*}{Ours}
 & $\texttt{I}$ (raw) & 60 & 100\% & 120 & 85.7\% & 100 & 100\% & 12 & 12\% & 292 & 73\% \\
 & $\texttt{I}$ (MLP) & 60 & 100\% & 120 & 85.7\% & 100 & 100\% & 12 & 12\% & 292 & 73\% \\
 & $\texttt{S}$ (raw) & 60 & 100\% & 120 & 85.7\% & 100 & 100\% & 12 & 12\% & 292 & 73\% \\
 & $\texttt{I}_{\mathrm{eff}}$ (raw) & 60 & 100\% & 119 & 85\% & 100 & 100\% & 12 & 12\% & 291 & 72.8\% \\
 & $\texttt{S}_{\mathrm{eff}}$ (raw) & 60 & 100\% & 119 & 85\% & 100 & 100\% & 12 & 12\% & 291 & 72.8\% \\
\bottomrule
\end{tabular}
}
\end{table*} 
    \Cref{tab:expressivity_results} compares the differentiated pairs of graphs using the full set of invariants, denoted by $\texttt{I}$, versus the methods listed in \citet{wang24a}.
In this comparison, invariants are the \emph{second} most expressive method after $N_2$, a $2$-hop neighbourhood counting method~\citep{papp22a}, outperforming \emph{all} GNN-based methods.
We note that by \emph{combining} invariants with existing methods, additional expressivity gains are possible.
For instance, local homomorphism counts can be combined with PPGN~\citep{maron19a} to differentiate up to $76.25\%$ of all pairs~\citep{jin24a}.\footnote{Care must be taken to perform the necessary ablations, though, since our analysis shows that homomorphism counts on their own are not sufficient to distinguish between large numbers of graphs on the BREC dataset.
    }
\begin{figure}
    \centering
    \includegraphics[width=1.0\textwidth]{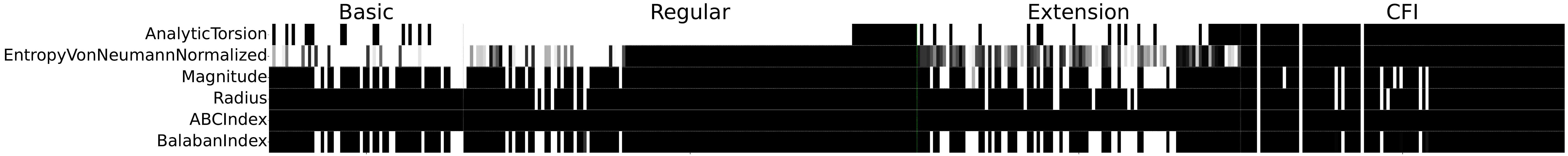}
    \caption{Heatmap that shows BREC graph pairs on the x-axis and invariants on the y-axis. The colour of a cell indicates the difference of an invariant value when computed for both graphs. Brighter cells indicate larger differences. The first four rows form an expressive subset.}
    \label{fig:heatmap_horizontal}
\end{figure} 
    A closer analysis of our full results indicates that expressivity does not require \emph{all} invariants.
Through greedy selection, i.e., by choosing invariants that differentiate the largest number of pairs on BREC, we observe that a subset of four invariants is sufficient to capture the \emph{same} expressivity as all combined invariants.
\Cref{fig:heatmap_horizontal} shows this \emph{expressive subset}, denoted by $\texttt{S}$, with the shading of each cell indicating the relative difference when evaluating the invariants.
The expressive subset comprises
\begin{inparaenum}[(i)]
        \item analytic torsion,
        \item mean commute time,
        \item magnitude, and
        \item the graph radius.
    \end{inparaenum}
Looking at the mean pairwise distance correlation between invariants over datasets in \Cref{fig:mean_min_corr_heatmap}, we see that this subset is \emph{not} strongly correlated.
We furthermore observe that the differentiated pairs of graphs substantially differ between invariants; the invariants are thus \emph{complementary}.
Notably, selecting invariants based on \emph{computational efficiency} yields a similar result. That is, removing computationally complex invariants~(cf.\ \Cref{tab:invariant_comp_time} for compute times), which yields the \emph{efficient} set $\texttt{I}_{\mathrm{eff}}$ and its expressive subset $\texttt{S}_{\mathrm{eff}}$, has negligible impact on expressivity~(more precisely, the difference is only a \emph{single} pair of graphs).
Hence, even for datasets with large graphs, which pose computational bottlenecks for the calculation of some invariants, we are able to choose subsets of invariants that are not redundant, maintain expressivity, and are efficient to compute.
More information about the datasets and the selected subsets of invariants is available in \Cref{app:regimes} and \Cref{tab:invariants_subsets}, respectively.

    \begin{tcolorbox}[title=In a nutshell]
      Invariants are expressive on their own. A subset of four invariants is sufficient to reach high expressivity on the BREC dataset. This result also holds when restricting the selection to invariants that remain computationally efficient for large graphs.
    \end{tcolorbox}

\subsection{Invariants highlight dataset heterogeneity}
\label{sec:Heterogeneity}

    Given their effectiveness as ``graph fingerprints,'' we next quantify to what extent invariants serve to \emph{structurally} characterize graph datasets \emph{as a whole}.
This analysis is complementary to prior work~\citep{Palowitch22a}, which shows that existing benchmark repositories such as the Open Graph Benchmark~\citep{hu20a} only cover a small fraction of the space of all possible graphs~(measured based on the clustering coefficient and the Gini coefficient of the degree distribution).
By contrast, our subsequent analysis aims to predict the dataset a graph belongs to based on its structural information only.
To this end, we treat dataset characterization as a ``meta-classification'' task: given a collection of datasets, we enumerate them and assign each graph the index of its source dataset as a label. We then sample graphs from each dataset and split them into training and test data. As input features, we use the \emph{concatenated invariants} of the graph, with the prediction target being dataset membership~(see \Cref{tab:xgboost_meta_clf} for XGBoost hyperparameters). Test accuracy then indicates to what extent the datasets are structurally separable.

    \paragraph{Pairwise confounder-free meta-classification.}
        An obstruction to meaningful measurements of dataset heterogeneity is that simple graph statistics such as the number of nodes already differentiate a large fraction of graphs due to differences in average graph size.
To measure dataset heterogeneity beyond these simple statistics, we remove these confounders for the following analysis.
To do so, we sample graphs in groups~(i.e., one per dataset), that all agree with respect to the following confounders: the number of nodes, the number of edges, and the second degree moment.
Hence, for any pair of datasets, this implies that we only sample pairs where these confounders coincide.
As a result, the three confounders, as well as all invariants derived from them~(such as the first degree moment or the graph density) are \emph{uninformative} for classification.
Due to the similar graph sizes in MoleculeNet~\citep{wu17a} datasets, we are able to sample graphs for an array of pairwise meta-classification experiments \emph{without} confounders.
Whenever possible, we sample up to $1{,}000$ samples per dataset~(matched graphs per pair in \Cref{tab:matched_pairs}), splitting all sampled graphs into a train/test set with an $80/20$ split.
\begin{table}[t]
\centering
\caption{{Pairwise meta-classification under matched sampling on MoleculeNet datasets.} We repeat sampling and classification with the full set of invariants for $5$ seeds. Cells show mean test accuracy and standard deviation. All datasets contribute graphs with matched confounder invariants, so confounders carry no signal. Pairs with fewer than $400$ matched graphs per dataset are excluded and marked with ``---''.}
\label{tab:confounder_matched_accuracy}
\small
\setlength{\tabcolsep}{4pt}
\begin{tabular}{l c c c c c c}
\toprule
 & BACE & BBBP & HIV & Lipo & PCBA & QM9 \\
\midrule
BACE & --- & 0.856 $\pm$ 0.028 & 0.885 $\pm$ 0.008 & 0.884 $\pm$ 0.024 & 0.887 $\pm$ 0.016 & --- \\
BBBP & 0.856 $\pm$ 0.028 & --- & 0.670 $\pm$ 0.014 & 0.660 $\pm$ 0.029 & 0.669 $\pm$ 0.023 & --- \\
HIV & 0.885 $\pm$ 0.008 & 0.670 $\pm$ 0.014 & --- & 0.687 $\pm$ 0.024 & 0.579 $\pm$ 0.018 & 0.718 $\pm$ 0.019 \\
Lipo & 0.884 $\pm$ 0.024 & 0.660 $\pm$ 0.029 & 0.687 $\pm$ 0.024 & --- & 0.665 $\pm$ 0.018 & --- \\
PCBA & 0.887 $\pm$ 0.016 & 0.669 $\pm$ 0.023 & 0.579 $\pm$ 0.018 & 0.665 $\pm$ 0.018 & --- & 0.738 $\pm$ 0.015 \\
QM9 & --- & --- & 0.718 $\pm$ 0.019 & --- & 0.738 $\pm$ 0.015 & --- \\
\bottomrule
\end{tabular}
\end{table}
 As \Cref{tab:confounder_matched_accuracy} shows, even graphs within the \emph{same} domain are structurally separable via graph invariants.

    \paragraph{Meta-classification across domains.}
        Moving to classification \emph{across} domains, we are unable to match all datasets based on confounders, so we sample $1{,}000$ graphs per dataset and otherwise follow the same evaluation procedure as above.
As \Cref{tab:meta_clf_acc_table} shows, invariants obtain an overall accuracy of $\approx 61\%$~(efficient set $\texttt{I}_{\mathrm{eff}}$) for classifying $26$ datasets from different domains~(social networks, small molecules, circuits, \dots), whereas confounders achieve an accuracy of $\approx 56\%$.
Further restricting the classification to one dataset per category~(cf.\ \cref{app:diverse_subset}), we achieve an accuracy of $\approx 99\%$.
The corresponding confusion matrix~(\Cref{fig:confusion_all}) shows that datasets \emph{within} the same domain tend to get confused with one another more than datasets from different domains.
Together with the results of the pairwise classification experiment, these results indicate a pronounced dataset shift, which remains prevalent in graphs from the same domain. Since this is known to pose an obstacle for generalization behaviour of graph models~\citep{Gui22a}, our analysis serves as a partial explanation as to why we still only observe limited generalization \emph{across} datasets and tasks in graph learning.

\begin{tcolorbox}[title=In a nutshell]
          Invariants can characterize existing graph datasets as a whole, highlighting a lack of homogeneous benchmark datasets that is conjectured to negatively impact graph learning research if not addressed~\citep{Bechler-Speicher25a}. \end{tcolorbox}
        
\subsection{Invariants capture structural relevance}
    \label{sec:relevance_of_structure}

    Finally, we carry out an extensive set of graph-level benchmarking operations, analysing the performance of invariants, features, or a mix of both.
All figures and tables in this section are based on \Cref{tab:full_table}, which shows the results for eight input configurations on $26$ datasets.
The aim of these experiments is to disentangle the task-specific relevance of
    \begin{inparaenum}[(i)]
        \item features,
        \item message-passing,
        \item expressivity, and
        \item global structure that is \emph{not} captured by expressivity alone.
    \end{inparaenum}
To this end, we evaluate four input configurations, chosen so that combinations of them reveal the contributing factors. Each configuration produces a single feature vector per graph, which is then supplied to XGBoost. For every dataset, we compute the set of tractable invariants specified in~\cref{app:regimes_and_invariant_sets}. The four configurations are:
\begin{compactenum}
        \item \texttt{sum}: sum-based feature representation
        \item \texttt{agg}: aggregation-based representation
        \item \texttt{I}: all invariants
        \item \texttt{S}: a minimal subset of expressive invariants~(cf.\ \Cref{sec:expressivity})
    \end{compactenum}
For datasets of the efficient regime~(\cref{app:regimes}), \texttt{I} and \texttt{S} refer to $\texttt{I}_{\mathrm{eff}}$ and $\texttt{S}_{\mathrm{eff}}$, respectively.
For the feature-based configurations \texttt{sum} and \texttt{agg}, we define an initial feature matrix that depends on whether a graph has edge features. If no such features exist, we set $X_\mathrm{init} = X$, where $X$ denotes the node feature matrix of the graph.
For graphs with edge features, we concatenate the sum of features of adjacent edges to each node, that is we define $X_\mathrm{init}=X \parallel B E$, where $B$ is the vertex--edge incidence matrix, $E$ the edge feature matrix, and $\parallel$ denotes concatenation.
This does \emph{not} increase the expressivity of message passing: for graphs with unit features, where expressivity is conventionally analysed, this operation only appends the node degree.
Finally, we calculate $\texttt{sum} \coloneqq  \sum_i (X_{\mathrm{init}})_i$ as the \emph{sum} of the feature matrix and  $\texttt{agg} \coloneqq ~ \parallel_i \sum_j (A^iX)_j$, i.e., the iterative \emph{concatenation} of neighbourhood features.
This calculation mimics message-passing as found in GIN, for instance, albeit without any nonlinearities or trainable aspects.
For the two invariant-based configurations, we concatenate invariants into a feature vector and make predictions based on combinations of these configurations to isolate the role of each factor~(i.e., features, message-passing, expressivity, or global structure).
    
    \paragraph{Experimental setup.}
        We keep hyperparameters and their ranges the same for all experiments~(see \Cref{app:xgboost-hyperparameters}) and  perform $150$ tuning iterations over the training data with early stopping, followed by a final fit with the \emph{best} set of hyperparameters.
The final fit of the model and its subsequent evaluation is repeated for $5$ seeds, over which we report mean and standard deviations.
We take care to work with official splits and according to the standard evaluation procedure whenever available~(see \Cref{tab:datasets_overview} for details).
For datasets with an official split, hyperparameters are tuned based on a $10$-fold cross-validation over combined train and validation data where feasible and based on validation score alone for datasets with more than $100{,}000$ samples.
For datasets with multiple targets, we tune one classifier per target.
For cross-validation, hyperparameters are tuned for each outer fold separately, with an inner $10$-fold cross-validation over the remaining folds.
We then fit and evaluate the best model for each outer fold and report the mean over the results.
This follows the suggested cross-validation protocol outlined by \citet{Errica2020A} and avoids overoptimistic reporting~\citep{OBray21a}.
We find that when trying to replicate the same approach with a conventional GNN like GIN~\citep{xu18a}, the model finishes \emph{at most} a single tuning iteration in the same time it takes to finish the tuning process for XGBoost.
This highlights the computational simplicity of our proposed graph invariant baseline predictor and we conclude that more complex models require larger parameter budgets, with their runtimes making a similar strategy infeasible.
We reiterate that our goal is \emph{not} to find the best set of hyperparameters to beat the state of the art, but to provide a reliable, reproducible, and simple performance baseline that can be readily applied to new datasets.
In fact, we believe that more complex models will benefit from such comparisons and should, in theory, be able to outperform it.

    \begin{figure}[tbp]
     \centering
     \begin{subfigure}[b]{0.495\textwidth}
         \centering
         \includegraphics[height=7.02cm]{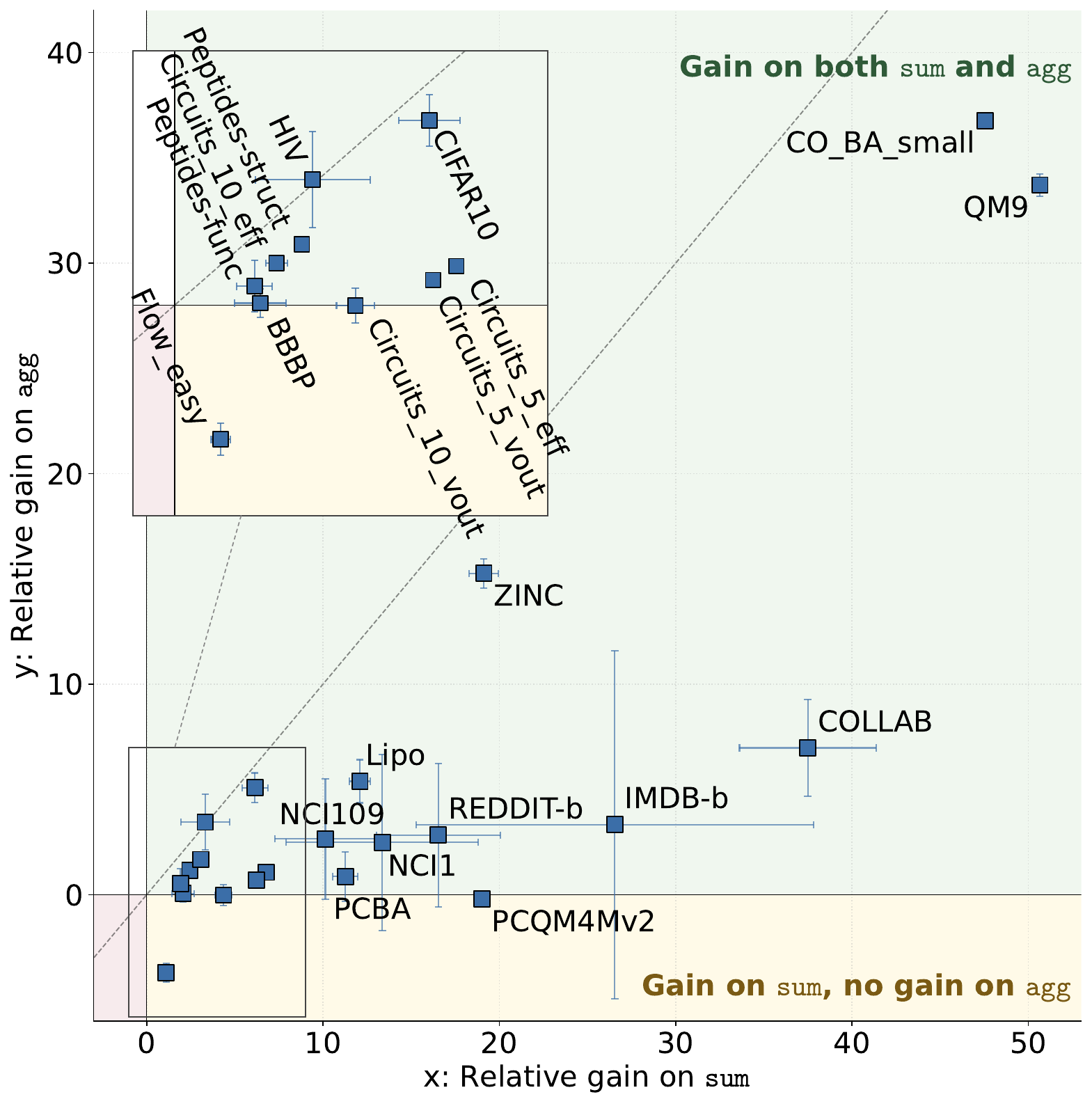}
         \caption{Appending \texttt{I}}
         \label{fig:image1}
     \end{subfigure}
\begin{subfigure}[b]{0.495\textwidth}
         \centering
         \includegraphics[height=7.02cm]{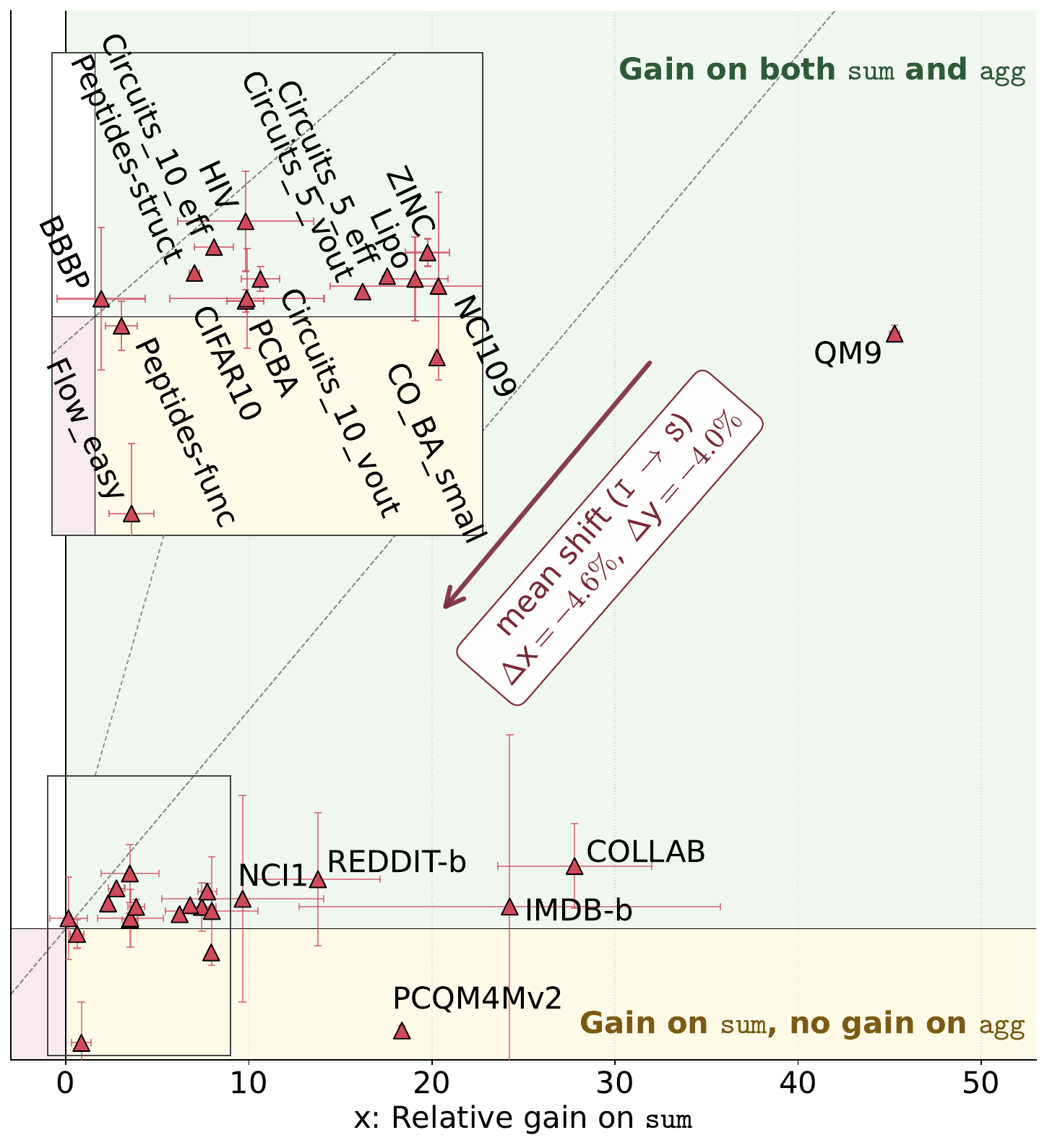}
         \caption{Appending \texttt{S}}
         \label{fig:image2}
     \end{subfigure}
     \caption{Relative gains from appending invariants. The x-axis shows relative gains over a \texttt{sum}-based classifier. The y-axis shows the relative gains over an \texttt{agg} based classifier. Zero on a single axis shows the performance of \texttt{sum} or \texttt{agg}, respectively.}
     \label{fig:relative_gains_combined}
\end{figure}         
    \paragraph{Insight: The importance of expressivity is questionable.}
\Cref{fig:relative_gains_combined} shows the relative performance gains obtained from adding invariants to \texttt{sum} and \texttt{agg}, respectively.
Whenever we observe a performance gain based on combining invariants and aggregated features, we see that the relative gain from adding invariants to summed features is larger. This indicates that our message-passing scheme already captures a substantial amount of the relevant structure---however, being upper-bounded by $1$-WL, this structure is not highly expressive.
When we reduce the number of invariants to the expressive subset of invariants $\texttt{S}$ or to the efficient subset $\texttt{S}_{\mathrm{eff}}$, both relative performance gains \emph{diminish} while expressivity stays the same.
In fact, there are even cases where the inverse is true. Predictive performance of the $1$-WL bounded \texttt{agg} does not increase to the same degree by concatenating invariants on all datasets. Examples are the BBBP and PCQM4Mv2 datasets~(cf.\ \Cref{tab:full_table}), where \texttt{agg} improves performance over \texttt{sum}, but concatenating invariants yields almost no further improvements.
This demonstrates that expressivity \emph{on its own} is not a good indicator of predictive performance: Assuming that settings like BREC are valid for measuring expressivity, a subset of invariants with the same expressivity as the full set of invariants nevertheless exhibits reduced predictive performance.

    \begin{wrapfigure}{r}{0.40\linewidth}
    \centering
    \vspace{-0.5\baselineskip}
    \includegraphics[width=0.95\linewidth]{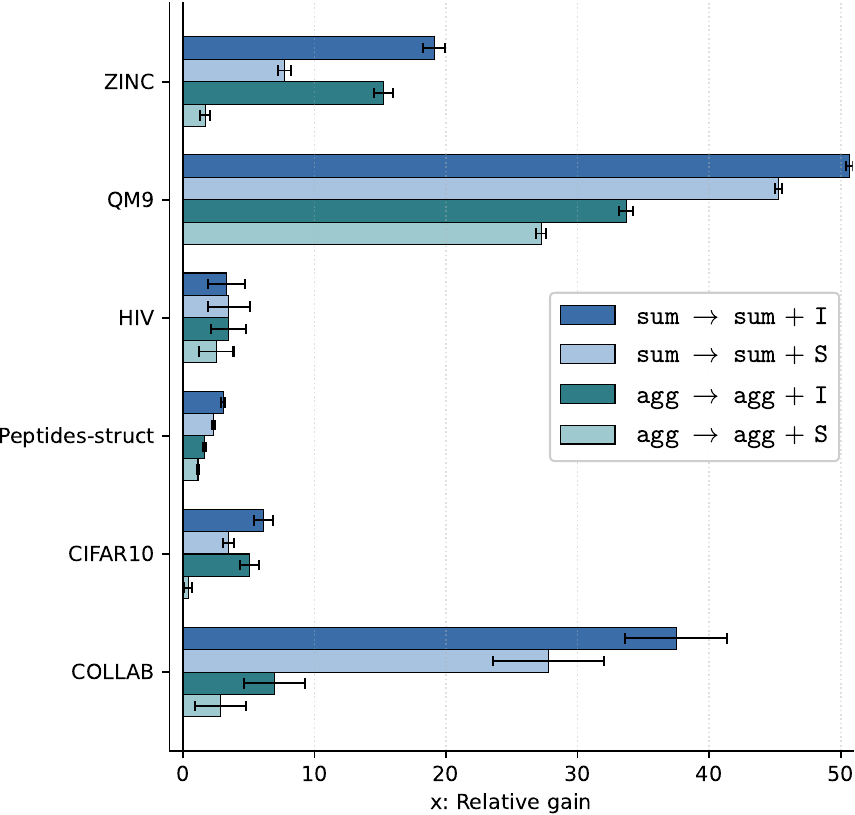}
    \caption{Datasets where we achieve consistent gains by appending invariants to the \texttt{sum} or \texttt{agg} representation.}
    \label{fig:consistent_structural_gains}
\end{wrapfigure} \paragraph{Insight: Invariants improve predictive performance over features alone.}
We also observe that in many cases, adding invariants to summed or aggregated features improves performance; cf.\ \Cref{fig:consistent_structural_gains}, in which invariants exhibit \emph{consistent} and \emph{substantial} relative performance gains over baselines~(both with the full set of invariants and with the subset).
These gains are particularly pronounced for the QM9 dataset, in which a combination of invariants and features exhibits competitive performance~(cf.\ \Cref{tab:qm9}; we will subsequently discuss this in more detail).
Notably, invariants on their own may perform poorly~(cf.\ ZINC in \Cref{tab:full_table}) and still capture complementary information to aggregated features.
Nevertheless, this analysis suggests that a combination of features and invariants results in a useful baseline model, which, as we demonstrated before, can be efficiently trained.
To further contextualize such relative performance gains, we next compare our graph invariant baseline model to results in literature.

\begin{figure}[tbp]
    \centering
    \includegraphics[width=\linewidth]{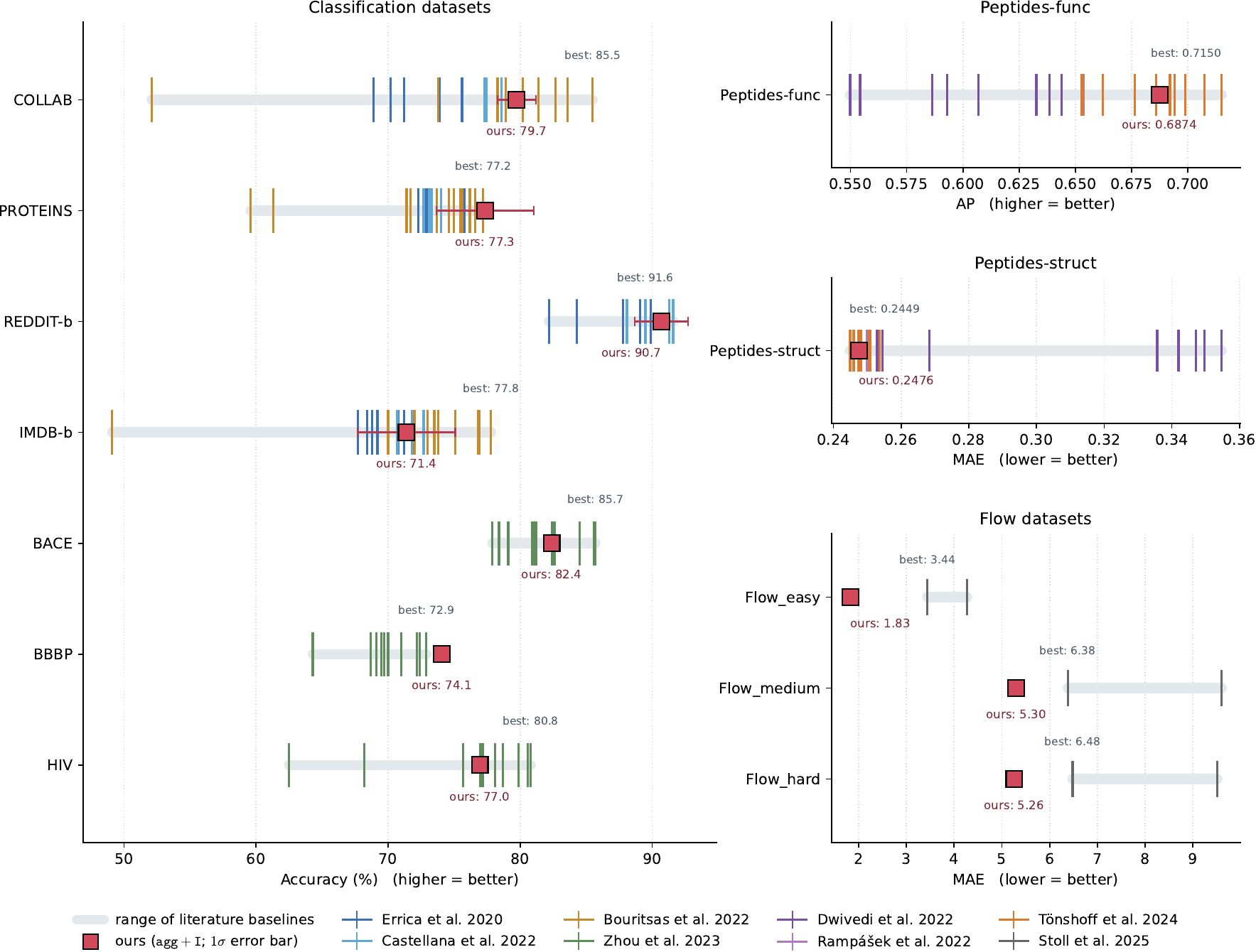}
    \caption{Datasets where $\texttt{agg}+\texttt{I}$ performs similarly or better than established methods. Tick colours correspond to source papers of reported performance values. Numeric values are in \Cref{tab:bm_tu} and \Cref{tab:bm_molecules} for classification datasets, in \Cref{tab:bm_peptides} for Peptides and in \Cref{tab:bm_flow} for flow datasets.}
    \label{fig:benchmark_comparison}
\end{figure}

    \paragraph{Insight: Tree ensembles are sufficient to solve some tasks.}
        We next compare our invariants baseline against results from the literature.
Such results usually emerge from a careful manual hyperparameter search, which tends to be more optimistic than what one might expect from a real application.
We therefore believe that our task-agnostic tuning procedure should not grant us any unfair advantages.
As \Cref{fig:benchmark_comparison} shows, on many datasets, invariant-based models perform on a par~(or better) with more powerful graph learning methods, including  transformer models and message-passing architectures.
When performing the same comparison for the \emph{minimal expressive subset}, with $\texttt{agg}+\texttt{S}$ in \Cref{fig:benchmark_comparison_subset}, we observe slightly reduced predictive performance, which nevertheless remains indistinguishable from that of more complex methods.
To substantiate this observation, we perform a \emph{multiple comparisons with the best} analysis~\citep[MCB]{hsu84a, Horrace20a}. This analysis indicates that $\texttt{agg}+\texttt{I}$ is not statistically significantly worse than the best method on the TU, MoleculeNet, and LRGB datasets, respectively~(see \Cref{app:mcb} for details).
Notice that for COLLAB, IMDB-b and REDDIT-b, we use the efficient invariant regime, but nevertheless obtain remarkably competitive predictive performance in comparison to more complex methods.
Most surprisingly, on novel datasets for solving max-flow problems~\citep{stoll26a}, our graph invariant baseline method even outperforms \emph{all} reported models by a substantial margin. 
A bootstrap confidence interval analysis further corroborates that these results are \emph{stable}~(see \Cref{tab:full_table_bootstrap,tab:qm9_bootstrap}).
We do \emph{not} interpret this as a sign to go back to feature engineering, but rather as a ``sanity check'' that provides information about the \emph{difficulty} of a task.
As a practical consideration, this nevertheless raises questions about the suitability of datasets, highly-parametrized models, and may ultimately indicate the need for additional hyperparameter tuning.
We therefore argue that standard baseline comparisons should be extended to include graph invariants in combination with features.

    \paragraph{Recommendation: Invariants should become a standard baseline.}
Given their empirical performance and simplicity, we believe that invariants constitute a natural minimum baseline, especially for datasets that do not provide features. Furthermore, understanding the relevance of each input configuration helps to find appropriate structural biases. For example, on BACE~\citep{wu17a}, a more complex model should be eschewed in favour of a smaller one since summed features alone lead to the same performance as aggregated features combined with invariants. In other cases, the choice of baseline heavily influences the \emph{perceived} performance of a method. For instance, homomorphism counts~\citep{jin24a} were compared to GNN-FiLM~\citep{brockschmidt20a}, with homomorphism counts clearly outperforming GNN-FiLM.
However, if we also consider EquiformerV2~\citep{yi-lun24a} as a comparison partner, performance differs roughly by a factor of $100$. The performance of $\texttt{agg}+\texttt{I}$, by contrast, is situated in-between~(i.e., \emph{better} than homomorphism count but worse than EquiformerV2) for most targets on QM9~(cf.\ \Cref{tab:qm9}).

        Notice that invariants do not always lead to high performance, as can be seen in \Cref{tab:full_table} for CIFAR10 or ``Electronic circuits,'' for instance. This is unsurprising for CIFAR10 since images are represented as pixel grid graphs. With all graphs then sharing the same underlying topology, purely structural invariants cannot distinguish them.

        Nevertheless, we believe that a complex model should be \emph{at least as good} as XGBoost with aggregated features, and ideally even better when adding invariants. This would indicate that such a model is able to extract structure and features in a meaningful manner.
We thus posit that simple invariant-based models should become a standard baseline to assess
\begin{inparaenum}[(i)]
            \item the relevance of structure in graph datasets~(by providing a simple structural baseline), as well as
            \item  the \emph{ability} of a model to make use of such structure~(by comparing model performance to invariant performance);
        \end{inparaenum}
as such, they may serve to address recent issues raised concerning graph machine learning benchmarks~\citep{Bechler-Speicher25a, Coupette25a}.
 \begin{tcolorbox}[title=In a nutshell]
    Invariants form a minimal yet competitive baseline, expose expressivity as measured on BREC as a poor predictor of task performance, and indicate when graph structure could matter for a task. They should thus be evaluated prior to employing more complex models in order to gauge the suitability of tasks and models early on.
\end{tcolorbox}

\section{Discussion}
    Graph invariants are simple, permutation-invariant, task-agnostic structural descriptors and serve as a versatile diagnostic lens for disentangling structure from features in graph benchmarks and as a
    surprisingly competitive baseline for graph learning. Returning to the three questions that frame this work, we found that
\begin{inparaenum}[(i)]
        \item invariants are more expressive than standard GNNs in the BREC sense,
        \item they characterize structural heterogeneity \emph{within} and \emph{across} benchmark
            datasets, and, somewhat alarmingly,
        \item simple invariant-based models are statistically on a par with transformer and message-passing baselines on the TU, MoleculeNet, and LRGB benchmarks, and outperform them on the novel Flow datasets.
    \end{inparaenum}

    \paragraph{Limitations.}
      Our analysis is restricted to graph-level tasks, and not all informative invariants
      are tractable on dense or large graphs. Yet, also with a computationally efficient set of invariants~(cf.\ \cref{app:regimes}), invariants retain their diagnostic and predictive merits.
Due to the nature of the datasets and the models, gains of models that combine features and invariants can also not directly be attributed to leveraging more structural information; it can only provide a hint as to whether certain aspects of the structure, captured by invariants, are conducive for a task.
Hence, the invariant sets we selected for this work should not be taken to be \emph{universally optimal}; our selection was necessarily driven by ease of implementation and we believe that finding ``the'' comprehensive set of invariants requires a different setup and may ultimately strongly depend on the task.

    \paragraph{Implications for graph benchmarks.} Two takeaways emerge from our experiments. First, the emphasis on \emph{expressivity} as a primary
        design objective deserves reconsideration: Invariant subsets with identical expressivity can yield different performance, and a $1$-WL-bounded aggregation scheme already captures most of the relevant structure on many datasets. Conversely, this means that a non-expressive model is not automatically a useless one.
Second, the structural shifts we observe across domains, and even \emph{within} domains,
        indicate that current benchmarks do not support learning the structural diversity necessary for 
        foundation models, offering a partial explanation for the lingering difficulty of graph pretraining~\citep{Bechler-Speicher25a, Liu25a}. Rather than proposing a return to hand-crafted features, we consider graph invariants to be a simple, useful diagnostic tool for demonstrating task complexity and model suitability.
That is, we see the performance of graph invariants to be a \emph{lower bound} that a model should easily surpass before pursuing more ambitious goals like generalization across domains. 
        
    \paragraph{Practical recommendations.}
        Following our extensive experiments, we make the following recommendations for practitioners: 
\textbf{(R1)} For any new graph dataset, fit the four configurations \texttt{sum}, \texttt{agg}, \texttt{I}~(or \texttt{S}), and their combinations with XGBoost. The resulting baseline helps pinpoint to what extent  features, aggregation, or global structure are required. At the same time, this baseline suggests the structural biases a model should incorporate.
\textbf{(R2)} When computationally intensive methods fail to substantially outperform invariant-based ones, the deficit likely reflects under-tuning or model--task misalignment~\citep{toenshoff2024where}; where they do outperform, the gap quantifies the learned interplay between structure and features beyond static structure.
\textbf{(R3)} If invariants provide no gain over features on a dataset, structure likely only plays a marginal role. This questions its suitability as a \emph{graph} dataset, regardless of its size.

\ifarXiv \subsection*{Acknowledgements}
    
    This work has received funding from the Swiss State Secretariat for
    Education, Research, and Innovation~(SERI).
The authors declare no competing interests. The funders had no role in
    the preparation of the manuscript or the decision to publish.
\fi

\clearpage

\subsection*{AI use statement}
    In this work, we used generative AI tools for skimming over experimental results to povide an initial asssesment, followed by a more detailed manual assessment. We have \emph{not} used generative AI tools for developing models or conceptual frameworks, proposing or refining hypotheses, designing or providing feedback on research methodology/experiments, implementing methods, assisting with translation, cleaning/reformatting datasets or supporting qualitative/thematic data analysis. The rest of the required disclosure tasks are not applicable to this work.

    Additionally, we used generative AI tools for drafting scientific figures, brainstorming, drafting code for already designed experiments, discovering research, searching for information, getting feedback on readability/consistency. We have not used generative AI for the other research subtasks recommended for disclosure. We have reviewed all AI-assisted work. LLM-generated code was verified line by line. Generated code as well as figures were extensively manually adjusted to fit our requirements. We take full responsibility for the final content of this work, including text, claims or artifacts produced with the aid of generative AI.

\subsection*{Reproducibility statement}

We have taken the following steps to ensure reproducibility.
All datasets are publicly available, \Cref{tab:datasets_overview} lists their sources and the split schemes we use.
All graph invariants are formally defined in \Cref{app:invariants}, and the invariant sets used in each regime are specified in \Cref{app:regimes}.
The hardware used for the XGBoost experiments is specified in \Cref{app:hyperparameters}.
The complete hyperparameter configurations and search spaces for all XGBoost and GIN models are listed in \Cref{app:hyperparameters}.
Procedures are described with each experiment: \Cref{sec:expressivity} follows \citet{wang24a}. Sampling, matching and the task setup for meta-classification is detailed in \Cref{sec:Heterogeneity}. Input configurations and evaluation protocols for XGBoost are detailed in \Cref{sec:relevance_of_structure}.
All results are reported with mean and standard deviation over $5$ random seeds, or over $10$ folds for datasets evaluated with nested cross validation. We additionally provide bootstrap confidence intervals~(\Cref{tab:full_table_bootstrap,tab:qm9_bootstrap}) and a statistical comparison with published results in \Cref{app:mcb}.
Literature values in \Cref{tab:bm_tu,tab:bm_molecules,tab:bm_peptides,tab:bm_flow} are taken from the papers cited in the same table.

\bibliography{main}
\bibliographystyle{abbrvnat}

\clearpage

\appendix
\raggedbottom

\counterwithin*{figure}{part}
\stepcounter{part}
\renewcommand{\thefigure}{S.\arabic{figure}}

\counterwithin*{table}{part}
\stepcounter{part}
\renewcommand{\thetable}{S.\arabic{table}}

\startcontents
\printcontents{}{1}{{\hrule
    \large\textbf{Appendix~(Supplementary Materials)}\vskip3pt\hrule\vskip5pt}
}
\vskip10pt
\hrule
\vskip5pt

\section{Invariants and methods}
    This section provides technical background for our analyses: the graph invariants used throughout the paper, and an introduction to GNNs.
    \newcolumntype{M}[1]{>{\raggedright\arraybackslash$\displaystyle}p{#1}<{$}}

\newcommand{\mathmode}{$}
\newcommand{\vdeg}{\deg}

\subsection{Invariants}
\label{app:invariants}
Here, we present the graph invariants used in the experiments, grouped into basic graph-theoretic invariants~(\Cref{tab:basic_invariants}), entropy-based invariants~(\Cref{tab:invariants_entropy}), geometric and topological invariants~(\Cref{tab:advanced_invariants}), and graph indices~(\Cref{tab:molecular_indices}). Before we start, we need to clarify the notation as well as recall and present some concepts.

We consider finite, undirected graphs without self-loops, denoted by $G \coloneq (V,E)$, where $V \coloneq \{1,\dots,n_V\}$ is the set of vertices and $E \subseteq \binom{V}{2}$ is the set of edges, with $n_E \coloneq |E|$. We denote the space of such graphs by $\mathcal{G}$. For a vertex $i\in V$, its neighbourhood is $N(i) \coloneq \{j\in V \mid \{i,j\}\in E\}$. Vertices and edges may carry features, denoted respectively by $\boldsymbol{v}_i \in \mathbb{R}^{d_V}$ and $\boldsymbol{e}_{ij} \in \mathbb{R}^{d_E}$.

The connectivity of $G$ is encoded by its adjacency matrix $\boldsymbol{A}\in\{0,1\}^{n_V\times n_V}$, defined by
\begin{equation*}
    (\boldsymbol{A})_{ij} \coloneq
    \begin{cases}
        1, & \text{if } j\in N(i),\\
        0, & \text{otherwise}.
    \end{cases}
\end{equation*}
The degree of a vertex $i$ is $\deg(i) \coloneq \sum_{j=1}^{n_V}(\boldsymbol{A})_{ij}$, and the shortest-path distance between two vertices $i,j\in V$ is denoted by $d(i,j)$. Equivalently, for $i\neq j$,
\begin{equation*}
    d(i,j) = \min\{k\geq 1 \mid (\boldsymbol{A}^k)_{ij}>0\},
\end{equation*}
with the convention that $d(i,j)=\infty$ if $i$ and $j$ lie in different connected components, and $d(i,i)=0$.

We denote by $\mathrm{spec}(\boldsymbol{M})$ the spectrum of a square matrix $\boldsymbol{M}$, that is, the set of its eigenvalues. The degree matrix $\boldsymbol{D}$ is defined by $(\boldsymbol{D})_{ii}\coloneq \deg(i)$, and the symmetrically normalized graph Laplacian is
\begin{equation*}
    \boldsymbol{L}_{\mathrm{sym}} \coloneq \boldsymbol{I} - \boldsymbol{D}^{-1/2}\boldsymbol{A}\boldsymbol{D}^{-1/2}.
\end{equation*}
If isolated vertices are present, we use the standard convention $(\boldsymbol{D}^{-1/2})_{ii}=0$ whenever $\deg(i)=0$.

The number of connected components of $G$ is denoted by $n_{\mathrm{connect}}$. The circuit rank, also called the cyclomatic number, is
\begin{equation*}
    \mathrm{rank}(G) \coloneq n_E - n_V + n_{\mathrm{connect}}.
\end{equation*}

Now, we present below the graph invariants from four families: basic graph-theoretic invariants, entropy-based invariants, geometric and topological invariants, and graph indices.

\begin{table}
    \renewcommand{\arraystretch}{1.8}
    \centering
    \caption{\textbf{Basic graph-theoretic invariants}. The number of unique triangles, i.e. three vertices forming a triangle, is denoted by $n_\mathrm{triangle}$, while the number of three vertices connected by at least two edges is $n_{\mathrm{triplet}}$. A graph is a \emph{tree} if it is connected with no \emph{cycles}. A graph $H = (V', E')$ is said to be a \emph{subgraph} of $G$, denoted by $H\subseteq G$, if $V'\subseteq V$, $E'\subseteq E$, and any edge in $E'$ connects two vertices in $V'$
    }
    \label{tab:basic_invariants}
    \vspace{5pt}
    \begin{tabular}{l M{6.0cm}}
        \toprule
        \textbf{Invariant} & \textbf{Definition} \\
        \midrule

        Number of vertices & 
        n_V \\

        Number of edges & 
       n_E \\
        
        Circuit rank & \mathrm{rank}(G)
         \\

        Diameter &
        \max\limits_{i,j \in V}d(i, j)\\

        Radius &
        \min\limits_{i \in V} \max\limits_{j \in V} d(i, j)\\

        Transitivity &  
        3 \frac{n_\mathrm{triangle}}{n_\mathrm{triplet}} \\

        Density &
        \frac{2 n_E}{n_V(n_V-1)} \\

        Eigenvalues of the normalized graph laplacian &
        \mathrm{spec}(\boldsymbol{L}_{\mathrm{sym}})
        \\

        Natural logarithm of the number of spanning trees &
        \ln \left|\{ H \subseteq G \mid H \text{ is a tree } \text{ and } V' = V \}\right|\\

        Geometric over arithmetic average of degree &
        \frac{\left(\prod\limits_{i \in V} \deg(i) \right)^{1/n_V}}{\frac{1}{n_V} \sum\limits_{i \in V}\deg(i)} \\

\bottomrule
    \end{tabular}
\end{table}

\begin{table}
    \renewcommand{\arraystretch}{1.8}
    \centering
    \caption{\textbf{Entropy-based invariants}. The function $\mathrm{order}:\{i,j\} \mapsto (\min(i,j), \max(i,j))$ converts an edge into an ordered pair. 
    The function $\mathrm{list}$ takes an edge set, applies $\mathrm{order}$ to every edge and then sorts them by lexicographical order to produce a sequence.
    The function $\mathrm{bytes}$ converts a sequence of edges into a byte sequence. The function $\mathrm{compress}$
    reduces the byte sequence into a smaller or equal byte sequence. The function $\mathrm{length}$
    returns the number of bytes in a given byte sequence.}
    \label{tab:invariants_entropy}
    \vspace{5pt}
    \begin{tabular}{l >{\raggedright\arraybackslash\mathmode}p{7.5cm}<{\mathmode}}
        \toprule
        \textbf{Invariant} & \textbf{Definition} \\
        \midrule

        Degree entropy & 
        -\sum_{k =1}^{n_V-1} \frac{|\{i\in V | \deg(i) = k\}|}{n_V} \log\left(\frac{|\{i\in V | \deg(i) = k\}|}{n_V}\right)\\

         von Neumann entropy &
        -\sum_{i=2}^{n_V} \frac{\lambda_i}{n_V} \log\left(\frac{\lambda_i}{n_V}\right), \quad \lambda_i\in \mathrm{spec}(\boldsymbol{L}_{\mathrm{sym}})\\

        Kolmogorov complexity & 
        \mathrm{length}(\mathrm{compress}(\mathrm{bytes}(E)))\\

        \bottomrule
    \end{tabular}
\end{table}

\begin{table}
    \renewcommand{\arraystretch}{1.8}
    \centering
    \caption{\textbf{Geometric and topological invariants}. 
    The square matrix $\boldsymbol{Z}_G(q)\in \mathbb{R}^{n_V \times n_V}$ is defined such that $(\boldsymbol{Z}_G(q))_{ij}\coloneq q^{d(i,j)}$, $ q\in \mathbb{R}$. For homomorphism counts, $\mathrm{hom}(F,G)$ denotes the set of all homomorphisms from a graph $F$ to $G$, and $\Omega^{\mathrm{con}}_{\le 5}$ denotes the family of all connected graphs with up to five vertices. For every $p>0$, we have the incidence matrix $\boldsymbol{B}_p$ (see \citep{alain2024} for more detail). For $p=0$, $\boldsymbol{B}_0=\boldsymbol{0}$. The $p$-Hodge Laplacian $\boldsymbol{L}_p$  is defined by $\boldsymbol{L}_p \coloneq \boldsymbol{B}_p^\top\boldsymbol{B}_p + \boldsymbol{B}_{p+1}\boldsymbol{B}_{p+1}^\top$.
    We denote Wasserstein distance of order $1$ by $W_1$ and $\mu_i$ is the probability measure of a lazy random walk associated with the vertex $i$. Both Forman-Ricci and Ollivier-Ricci curvature form distributions over the edges. We use either the empiric mean, variance, skewness or kurtosis of these distributions as an invariant. The commute time is defined for every pair $(i,j)\in V\times V$, with $\boldsymbol{L}^{\dag}$ the Moore--Penrose pseudoinverse of the unnormalized Laplacian $\boldsymbol{L}\coloneq\boldsymbol{D}-\boldsymbol{A}$ and $\mathrm{vol}(V)=2n_E$; we use the empirical mean and maximum over $V\times V$ as invariants. For the neighbourhood power trace, $\boldsymbol{A}|_{N(i)}$ denotes the principal submatrix of $\boldsymbol{A}$ indexed by the neighbourhood $N(i)$ of vertex $i$, following the original definition of \cite{duda2024a}. As a variant, we also consider the closed neighbourhood $N[i]\coloneq N(i)\cup\{i\}$ in place of $N(i)$.}
    \label{tab:advanced_invariants}
\vspace{5pt}
    \begin{tabular}{l >{\raggedright\arraybackslash\mathmode}p{6.5cm}<{\mathmode}}
        \toprule
        \textbf{Invariant} & \textbf{Definition} \\
        \midrule

        Magnitude, \cite{LEINSTER_2017} & 
        \sum_{i\in V}\sum_{j\in V}(\boldsymbol{Z}(q)^{-1})_{ij}\\
        
        Analytic torsion, \cite{knill2022analytictorsiongraphs} & 
        \prod_{p=0}^{P}\mathrm{det}(\mathbf{L}_p)^{p(-1)^{p+1}} \\
        
        Homomorphism counts, \cite{jin24a} &
        |\mathrm{hom}(F, G)|, \quad F\in\Omega^{\mathrm{con}}_{\le 5}\\

        Forman--Ricci curvature & 
        4 - \left( \deg(i) + \deg(j) \right)\\

        Ollivier--Ricci curvature & 
        1 - \frac{W_1(\mu_i, \mu_j)}{d(i,j)}\\

        Commute time &
        \mathrm{vol}(V)\bigl((\boldsymbol{L}^{\dag})_{ii} + (\boldsymbol{L}^{\dag})_{jj} - 2(\boldsymbol{L}^{\dag})_{ij}\bigr)\\

        Neighbourhood power trace, \cite{duda2024a} &
        \sum_{i\in V}\mathrm{tr}\bigl((\boldsymbol{A}|_{N(i)})^{p}\bigr), \quad p\in\{4,8\}\\
        
        \bottomrule
    \end{tabular}
\end{table}

\begin{table}
        \renewcommand{\arraystretch}{1.8}
        \centering
        \caption{\textbf{Graph indices}. These graph invariants are commonly used in molecular applications. For an edge $\{i,j\}\in E$, the number of vertices that are closer to $i$ than to $j$ is $n_i({\{i,j\}}) \coloneq |\{ k \in V \mid d(k, i) < d(k, j) \}|$.}
        \label{tab:molecular_indices}
        \vspace{5pt}
        \begin{tabular}{l >{\raggedright\arraybackslash\mathmode}p{7.8cm}<{\mathmode}}
            \toprule
            \textbf{Invariant} & \textbf{Definition} \\
            \midrule
    
            Wiener index & 
            \frac{1}{2}\sum_{i\in V}\sum_{j\in V}d(i, j) \\
            
            Randić index & 
            \sum_{\{i, j\}\in E}\frac{1}{\sqrt{\vdeg(i)\vdeg(j)}} \\
            
            General Randić index & 
            \sum_{\{i, j\}\in E}\big(\mathrm{deg}(i)\vdeg(j)\big)^{c}, \quad c\in\mathbb{R} \\
            
            Atom bond connectivity & 
            \sum_{\{i,j\}\in E}\sqrt{\frac{\vdeg(i) + \vdeg(j) - 2}{\vdeg(i)\vdeg(j)}} \\
            
            Geometric Arithmetic index & 
            \sum_{\{i,j\}\in E}2\frac{\sqrt{\vdeg(i)\vdeg(j)}}{\vdeg(i)+\vdeg(j)} \\
            
            Hyper-Wiener index & 
            \frac{1}{2}\sum_{i \in V}\sum_{j \in V}\left(d(i, j)+d(i,j)^2\right) \\
            
            Estrada index & 
            \sum_{\lambda_i\in\mathrm{spec}(\boldsymbol{A})} \exp(\lambda_i) \\
            
            First Zagreb index & 
            \sum_{i\in V}\vdeg(i)^2 \\

            Second Zagreb index & 
            \sum_{\{i, j\}\in E}\vdeg(i)\vdeg(j) \\
            
            Schultz index & 
            \frac{1}{2}\sum_{i\in V}\sum_{j\in V}d(i,j)\big(\vdeg(i) + \vdeg(j)\big) \\
            
            Gutman index & 
            \frac{1}{2}\sum_{i\in V} \sum_{j\in V}d(i,j)\vdeg(i)\vdeg(j) \\
            
            Szeged index & 
            \sum_{\{i,j\}\in E}n_i(\{i,j\})n_j(\{i,j\})  \\
            
            Forgotten index & 
            \sum_{i\in V}\vdeg(i)^3 \\

            Balaban index & 
            \frac{m}{\mathrm{rank}(G)+1} \sum_{\{i,j\}\in E} \big(\sum_{k\in V}d(i,k)\sum_{l\in V} d(j,l)\big)^{-1/2}\\
            
            \bottomrule
        \end{tabular}
    \end{table}     \clearpage

    \subsection{Methods}

\paragraph{Graph neural networks.}

In this subsection, we briefly present graph neural networks. Graphs contain structural information that is often essential for machine learning tasks, and the central question is how to incorporate this structure into a learning system.

Modern graph neural networks (GNNs) commonly rely on the message-passing (MP) paradigm, which consists of two steps: \emph{aggregation} and \emph{update}. For simplicity, we discuss only vertex features; edge features can be handled similarly. The algorithm proceeds layer by layer. At layer $k=0$, each vertex $i$ is initialized with its feature vector,
\begin{equation*}
    \boldsymbol{h}_i^{(0)} \coloneq \boldsymbol{v}_i .
\end{equation*}
For $k\geq 0$, the hidden representations of neighbouring vertices are aggregated into a message
\begin{equation*}
    \boldsymbol{m}^{(k+1)}_i
    \coloneq
    \mathrm{aggregate}\left(
    \left[\boldsymbol{h}^{(k)}_j \mid j\in \mathcal{N}(i)\right]
    \right),
\end{equation*}
where $\mathrm{aggregate}$ is permutation-invariant and $[\cdot]$ denotes a multiset. The vertex representation is then updated by
\begin{equation*}
    \boldsymbol{h}^{(k+1)}_i
    \coloneq
    \mathrm{update}\left(
    \boldsymbol{h}^{(k)}_i,
    \boldsymbol{m}_i^{(k+1)}
    \right).
\end{equation*}
These operations are applied in parallel to all vertices. Typically, aggregation uses the adjacency matrix $\boldsymbol{A}$ to collect information from neighbouring vertices, while the update step is implemented by learnable transformations and nonlinear activation functions.

In short, message passing iteratively incorporates local graph structure into vertex representations. However, MP-based GNNs are at most as expressive as the \emph{$1$-Weisfeiler--Lehman} test for graph isomorphism~\citep{xu18a}. Consequently, globally different graphs may be indistinguishable from the local perspective of such models, motivating the use of global inductive biases.     \clearpage

\section{Dataset regimes and invariant sets}
    \label{app:regimes_and_invariant_sets}
    This appendix shows the datasets used in our experiments as well as the invariant regimes and sets based on computational constraints.
    \begin{table}[H]
\centering
\caption{Statistics of investigated Graph Datasets. An asterisk $^*$ denotes splits were provided from an official source, otherwise splitting was done locally. Statistics were all computed locally. For datasets without node features, we introduced constant unit node features. For PCQM4Mv2 we are only using training and validation split data, since test data is not publicly available. Category denotes the application domain, or the task family for the synthetic benchmarks of \citet{stoll26a}.}
\label{tab:datasets_overview}
\resizebox{\textwidth}{!}{
\begin{tabular}{llcccccccccr}
\toprule
\textbf{Category} & \textbf{Name} & \textbf{Node Feat.} & \textbf{Edge Feat.} & \textbf{Targ.} & \textbf{Split Scheme} & \textbf{Task Type} & \textbf{Metric} & \textbf{\# Graphs} & \textbf{Avg. $|V|$} & \textbf{Avg. $|E|$} & \textbf{Avg. Deg.} \\ \midrule
Small molecules & BACE \cite{wu17a, Fey19a} & Yes & Yes & 1 & Scaffold & Binary Class. & ROC-AUC & 1,513 & 34.1 & 36.9 & 2.16 \\
Small molecules & BBBP \cite{wu17a, Fey19a} & Yes & Yes & 1 & Scaffold & Binary Class. & ROC-AUC & 2,039 & 24.1 & 26.0 & 2.16 \\
Small molecules & HIV \cite{hu20a}& Yes & Yes & 1 & Scaffold$^*$ & Binary Class. & ROC-AUC & 41,127 & 25.5 & 27.5 & 2.15 \\
Small molecules & ZINC (12k) \cite{dwivedi23a, Fey19a} & Yes & Yes & 1 & Random$^*$ & Regression & MAE & 12,000 & 23.2 & 24.9 & 2.15 \\
Small molecules & Lipo \cite{wu17a, Fey19a} & Yes & Yes & 1 & Scaffold & Regression & MSE & 4,200 & 27.0 & 24.5 & 2.18 \\
Small molecules & PCQM4Mv2 \cite{hu20a, Fey19a} & Yes & Yes & 1 & PubChem ID$^*$ & Regression & MAE & $\sim$3.6 million & 14.1 & 14.5 & 2.06 \\
Small molecules & PCBA \cite{hu20a}& Yes & Yes & 128 & Scaffold$^*$ & Multi-label & AP & 437,929 & 26.0 & 28.1 & 2.16 \\
Small molecules & QM9 \cite{wu17a, Fey19a} & Yes & Yes & 12 & Random & Regression & MAE & 130,831 & 18.0 & 18.7 & 2.07 \\ \midrule
Small molecules & NCI1 \cite{Morris20a, Fey19a} & Yes & No & 1 & 10-fold CV & Binary Class. & Accuracy & 4,110 & 29.9 & 32.3 & 2.16 \\
Small molecules & NCI109 \cite{Morris20a, Fey19a} & Yes & No & 1 & 10-fold CV & Binary Class. & Accuracy & 4,127 & 29.7 & 32.1 & 2.17 \\
Peptides & Peptides-func \cite{dwivedi22a, Fey19a} & Yes & Yes & 10 & Stratified$^*$ & Multi-label & AP & 15,535 & 150.9 & 153.7 & 2.04 \\
Peptides & Peptides-struct \cite{dwivedi22a, Fey19a} & Yes & Yes & 11 & Stratified$^*$ & Regression & MAE & 15,535 & 150.9 & 153.7 & 2.04 \\ \midrule
Bioinformatics & PROTEINS \cite{Morris20a, Fey19a} & Yes & No & 1 & 10-fold CV & Binary Class. & Accuracy & 1,113 & 39.1 & 72.8 & 3.73 \\
Comp. vision & CIFAR10 \cite{dwivedi23a, Fey19a}& Yes & Yes & 10 & Stratified$^*$ & Multi-class & Accuracy & 60,000 & 117.6 & 470.5 & 8.0 \\ \midrule
Social networks & COLLAB \cite{Morris20a, Fey19a} & No & No & 3 & 10-fold CV & Multi-class & Accuracy & 5,000 & 74.5 & 2457.2 & 65.97 \\
Social networks & IMDB-b \cite{Morris20a, Fey19a} & No & No & 1 & 10-fold CV & Binary Class. & Accuracy & 1,000 & 19.8 & 96.5 & 9.76 \\
Social networks & REDDIT-b \cite{Morris20a, Fey19a} & No & No & 1 & 10-fold CV & Binary Class. & Accuracy & 2,000 & 429.6 & 497.8 & 2.32 \\ \midrule
El. circuits & Circuits\_5\_eff \cite{stoll26a} & Yes & No & 1 & Random$^*$ & Regression & RSE & 334,410 & 10.1 & 10.0 & 1.98 \\
El. circuits & Circuits\_5\_vout \cite{stoll26a} & Yes & No & 1 & Random$^*$ & Regression & RSE & 334,419 & 10.1 & 10.0 & 1.98 \\
El. circuits & Circuits\_7\_eff \cite{stoll26a} & Yes & No & 1 & Random$^*$ & Regression & RSE & 1,295 & 12.1 & 14.0 & 2.32 \\
El. circuits & Circuits\_10\_eff \cite{stoll26a} & Yes & No & 1 & Random$^*$ & Regression & RSE & 4,630 & 16.0 & 20.0 & 2.51 \\
El. circuits & Circuits\_10\_vout \cite{stoll26a} & Yes & No & 1 & Random$^*$ & Regression & RSE & 4,630 & 16.0 & 40.0 & 2.51 \\ \midrule
Comb. optimization & CO\_BA\_small \cite{stoll26a} & No & No & 1 & Not found$^*$ & Binary Class. & MAE & 50,000 & 249.9 & 495.8 & 3.97 \\
Alg. reasoning & Flow\_easy \cite{stoll26a} & Yes & Yes & 1 & Not found$^*$ & Regression & MAE & 1,010,000 & 16.5 & 40.2 & 4.88 \\
Alg. reasoning & Flow\_medium \cite{stoll26a} & Yes & Yes & 1 & Not found$^*$ & Regression & MAE & 1,010,000 & 16.5 & 24.2 & 2.93 \\
Alg. reasoning & Flow\_hard \cite{stoll26a} & Yes & Yes & 1 & Not found$^*$ & Regression & MAE & 1,010,000 & 16.5 & 24.2 & 2.93 \\ \bottomrule
\end{tabular}
}
\end{table} 
    \paragraph{Computation time.}
        \label{app:computation_time}
        \Cref{tab:invariant_comp_time} reports the mean wall-clock time per graph of each invariant on four datasets, ranging from small molecules to large pixel-grid graphs.
These measurements are the basis for the regimes defined below.
        \begin{table}[H]
\centering
\small
\setlength{\tabcolsep}{4pt}
\resizebox{\textwidth}{!}{
\begin{tabular}{l r r r r r}
\toprule
 & \multicolumn{5}{c}{\textit{Time per graph (ms)}} \\
\cmidrule(lr){2-6}
Invariant & \textbf{PCQM4Mv2} & \textbf{BACE} & \textbf{Peptides-struct} & \textbf{CIFAR10} & \textbf{COLLAB} \\
\midrule
Analytic torsion & 0.40 & 0.84 & 4.4 & 129 & --- \\
Atom bond connectivity & 0.15 & 0.37 & 1.6 & 7.5 & 404 \\
Balaban index & 0.17 & 0.40 & 7.1 & 9.8 & 14 \\
Circuit rank & 0.045 & 0.099 & 0.43 & 0.63 & 1.8 \\
Commute time & 0.35 & 0.59 & 4.2 & 4.4 & 12 \\
Degree entropy & 0.033 & 0.062 & 0.23 & 0.34 & 1.4 \\
Density & 0.022 & 0.049 & 0.21 & 0.31 & 1.4 \\
Diameter & 0.31 & 1.6 & 40 & 33 & 75 \\
Eigenvalues of the normalized graph Laplacian & 0.43 & 0.60 & 2.4 & 2.5 & 11 \\
Estrada index & 0.12 & 0.28 & 2.7 & 2.5 & --- \\
Forgotten index & 0.032 & 0.071 & 0.31 & 0.40 & 1.5 \\
Forman--Ricci curvature & 0.13 & 0.32 & 1.3 & 12 & 1270 \\
General Randi\'{c} index & 0.098 & 0.24 & 1.1 & 4.3 & 207 \\
Geometric Arithmetic index & 0.13 & 0.32 & 1.4 & 7.3 & 407 \\
Geometric over arithmetic average of degree & 0.20 & 0.28 & 0.64 & 0.89 & 3.2 \\
Gutman index & 0.14 & 0.32 & 6.9 & 9.4 & 9.5 \\
Homomorphism counts ($\Omega^{\mathrm{con}}_{\le 5}$) & 76 & 82 & 83 & 687 & --- \\
Hyper-Wiener index & 0.30 & 1.7 & 42 & 35 & 78 \\
Magnitude & 0.14 & 0.31 & 7.0 & 9.5 & 8.2 \\
Neighbourhood power trace & 0.84 & 3.1 & 52 & 55 & 1492 \\
Number of edges & 0.022 & 0.049 & 0.21 & 0.32 & 1.4 \\
Natural logarithm of the number of spanning trees & 0.35 & 0.48 & 1.3 & 2.1 & 9.0 \\
Number of vertices & 0.0019 & 0.0025 & 0.0055 & 0.0077 & 0.019 \\
Ollivier--Ricci curvature & 1.3 & 3.1 & 18 & 51 & 29604 \\
Radius & 0.30 & 1.6 & 40 & 33 & 76 \\
Randi\'{c} index & 0.10 & 0.25 & 1.1 & 4.3 & 210 \\
Schultz index & 0.18 & 0.58 & 14 & 13 & 13 \\
Szeged index & 0.41 & 2.3 & 56 & 72 & 518 \\
Transitivity & 0.097 & 0.22 & 0.94 & 4.1 & 209 \\
Wiener index & 0.11 & 0.26 & 6.6 & 9.1 & 7.9 \\
Zagreb indices & 0.11 & 0.26 & 1.2 & 4.4 & 209 \\
von Neumann entropy & 0.37 & 0.49 & 2.1 & 2.0 & 8.3 \\
\midrule
\textit{Total} & 84 & 103 & 400 & 1206 & 34863 \\
\bottomrule
\end{tabular}
}
\caption{\textbf{Computation time of the invariants.} Mean wall-clock time per graph (ms) on a single CPU core. Means are taken over $5{,}000$ randomly sampled graphs for PCQM4Mv2, Peptides-struct and CIFAR10, and over all graphs for BACE and COLLAB. Entries marked ``---'' were not computable on this dataset.}
\label{tab:invariant_comp_time}
\end{table}
 
\begin{figure}[H]
    \centering
    \includegraphics[width=1.0\textwidth]{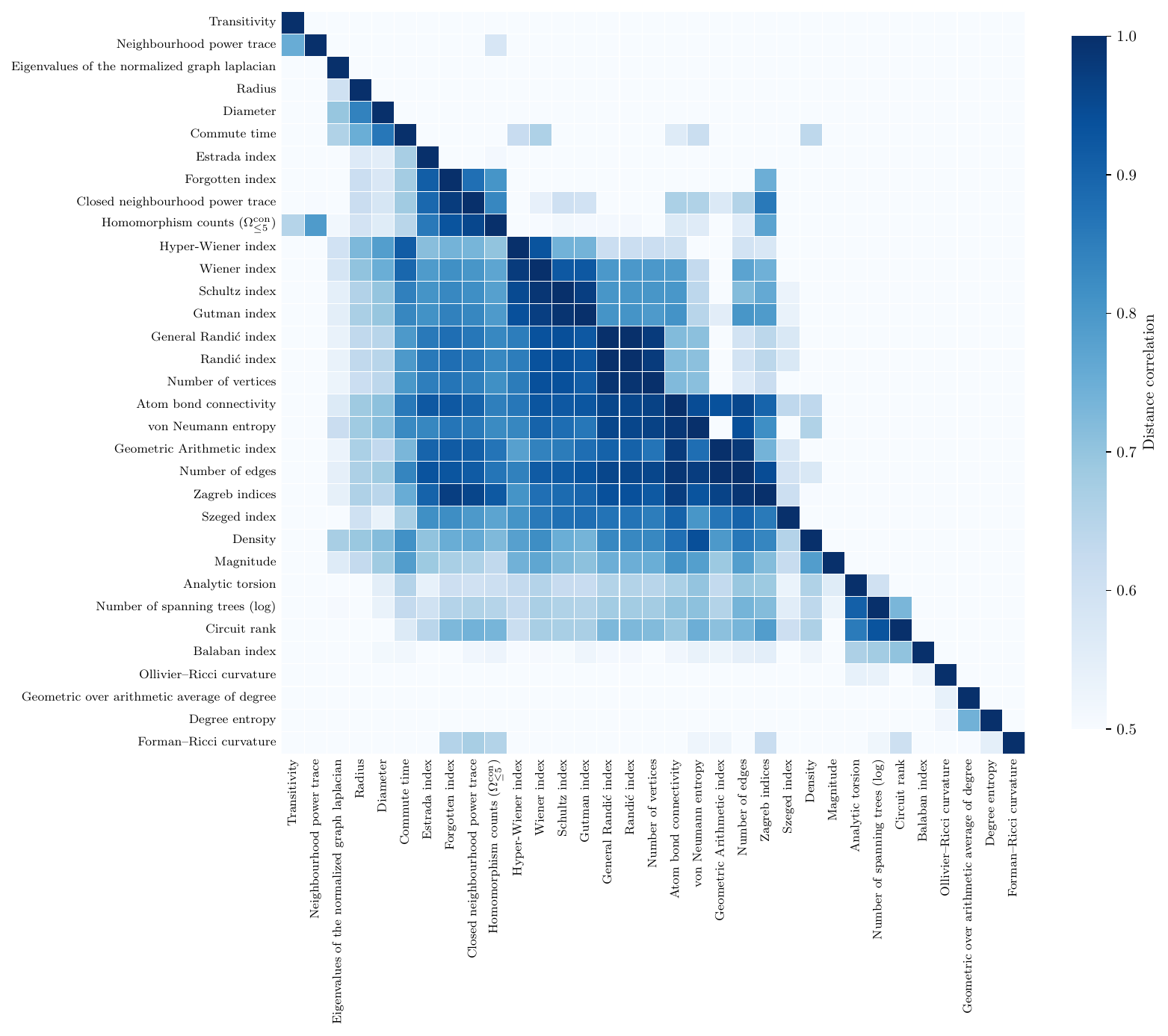}
    \caption{Distance correlation over the full regime. The lower triangle shows the mean, while the upper triangle shows the minimum of the distance correlation over the full regime.}
    \label{fig:mean_min_corr_heatmap}
\end{figure}

    \paragraph{Dataset regimes and invariant sets.}
        \label{app:regimes}
        We distinguish two regimes of datasets, summarized in \Cref{tab:invariants_subsets}.
In the \emph{full regime}, comprising all categories of \Cref{tab:datasets_overview} except Social networks and Comp.\ vision, we compute the complete set $\texttt{I}$ of invariants introduced in \cref{app:invariants}.
The \emph{efficient regime} comprises the remaining datasets COLLAB, IMDB-b, REDDIT-b, and CIFAR10.
COLLAB, CIFAR10, and IMDB-b have the highest average node degree in our collection, which makes the most expensive invariants intractable; for REDDIT-b, we suspect that a small number of unusually complex graphs prevent a successful computation.
For these datasets, we use the pruned set $\texttt{I}_{\mathrm{eff}}$, which omits the prohibitively expensive or numerically unstable analytic torsion, homomorphism counts, and Estrada index.
In both regimes, we additionally consider an \emph{expressive} subset~($\texttt{S}$ or $\texttt{S}_{\mathrm{eff}}$), i.e., the smallest subset of invariants that retains full expressivity on BREC~(\Cref{sec:expressivity}).
For all sets, the magnitude is evaluated at $q = e^{-0.42}$, selected by maximizing expressivity on BREC over a grid of candidate values.
        \begin{table}[H]
    \renewcommand{\arraystretch}{1.3}
    \centering
    \caption{\textbf{Invariant sets.} The full set $\texttt{I}$ contains all invariants introduced in~\Cref{app:invariants}, with its efficient-regime counterpart $\texttt{I}_{\mathrm{eff}}$. The expressive subsets $\texttt{S}$ and $\texttt{S}_{\mathrm{eff}}$ contain a small number of invariants identified as highly expressive on BREC~(\Cref{sec:expressivity}). Datasets are given by their category in \Cref{tab:datasets_overview}; each category comprises all of its datasets.}
    \label{tab:invariants_subsets}
    \resizebox{\textwidth}{!}{
    \begin{tabular}{l l l l}
        \toprule
        \textbf{Regime} & \textbf{Datasets} & \textbf{Set} & \textbf{Invariants} \\
        \midrule
        \multirow{2}{*}{Full}
            & Small molecules, Peptides, & $\texttt{I}$ & All invariants from the invariants section \\
            & Bioinformatics, El. circuits, & $\texttt{S}$ & Analytic torsion, mean commute time, magnitude ($q=e^{-0.42}$), \\
            & Comb. optimization, Alg. reasoning & & graph radius \\
        \midrule
        \multirow{2}{*}{Efficient}
            & Social networks, & $\texttt{I}_{\mathrm{eff}}$ & All invariants from the invariants section, except: analytic torsion, \\
            & Comp. vision & & homomorphism counts, and the Estrada index \\
            & & $\texttt{S}_{\mathrm{eff}}$ & Algebraic connectivity (second-smallest Laplacian eigenvalue), \\
            & & & mean Ollivier--Ricci curvature, magnitude ($q=e^{-0.42}$), \\
            & & & closed-neighbourhood power trace with $p=8$, graph radius \\
        \bottomrule
    \end{tabular}
    }
\end{table} 
    \paragraph{Redundancy between invariants.}
        \label{app:invariant_redundancy}
        \Cref{fig:mean_min_corr_heatmap} shows the pairwise distance correlation between invariants, computed over the datasets of the full regime.
Distance correlation measures the dependence between two random vectors and vanishes if and only if they are independent~\citep{Székely07a}.
It is thus preferable to the Pearson correlation in our setting, since invariants may depend on each other in a non-linear manner, which a purely linear measure would not detect.
We compute it per dataset with the \texttt{dcor} package~\citep{ramos23a} and report the mean and the minimum across datasets.
While some invariants are strongly correlated, those of the expressive subset $\texttt{S}$ are not, indicating that they capture complementary structural information.
\clearpage
\section{Hyperparameters}
    \label{app:hyperparameters}
    This appendix details the hyperparameters used for the XGBoost classifiers in our heterogeneity experiments (\cref{sec:Heterogeneity}) and downstream experiments (\cref{sec:relevance_of_structure}), and for the shared GIN graph-encoder used in the multi-task experiments in \cref{sec:Multi-Task Performance}.
In terms of hardware, our training used $6$ NVIDIA RTX 3080 and $4$ NVIDIA L40s GPUs.
    
    \paragraph{XGBoost.}
        \label{app:xgboost-hyperparameters}    
        We use XGBoost~\cite{chen16a} with histogram-based tree construction. For the downstream experiments~(\cref{sec:relevance_of_structure}), we train on GPU; the base configuration is given in Table~\ref{tab:xgboost_base}, and the hyperparameter search space explored by the Optuna~\cite{akiba19a} Bayesian sampler is listed in Table~\ref{tab:xgboost_search_space}. The number of boosting rounds is not tuned directly; instead, training stops early once the validation metric has not improved for $8$ rounds, and evaluation uses the best number of rounds. We run $150$ tuning iterations.
        \begin{table}[H]
    \renewcommand{\arraystretch}{1.3}
    \centering
    \caption{\textbf{Architectural XGBoost parameters} used in the downstream experiments.}
    \label{tab:xgboost_base}
    \begin{tabular}{l l}
        \toprule
        \textbf{Hyperparameter} & \textbf{Value} \\
        \midrule
        \texttt{device}             & \texttt{cuda} \\
        \texttt{tree\_method}       & \texttt{hist} \\
        \texttt{grow\_policy}       & \texttt{lossguide} \\
        Early-stopping patience     & 8 \\
        \bottomrule
    \end{tabular}
\end{table}         \begin{table}[H]
    \renewcommand{\arraystretch}{1.3}
    \centering
    \caption{\textbf{XGBoost hyperparameter search space} explored by the Optuna sampler.}
    \label{tab:xgboost_search_space}
    \begin{tabularx}{\textwidth}{l X l}
        \toprule
        \textbf{Hyperparameter} & \textbf{Range} & \textbf{Sampling} \\
        \midrule
        \texttt{learning\_rate}    & $[5\!\times\!10^{-3},\ 1.0]$ & log-uniform \\
        \texttt{max\_leaves}       & $[5,\ 1000]$                 & log-uniform (integer) \\
        \texttt{min\_child\_weight}& $[10^{-3},\ 200]$            & log-uniform \\
        \texttt{subsample}         & $[0.1,\ 1.0]$                & uniform \\
        \texttt{colsample\_bytree} & $[0.1,\ 1.0]$                & uniform \\
        \texttt{reg\_lambda}       & $[10^{-5},\ 100]$            & log-uniform \\
        \texttt{reg\_alpha}        & $[10^{-5},\ 100]$            & log-uniform \\
        \bottomrule
    \end{tabularx}
\end{table} 
    \paragraph{XGBoost for meta-classification.}
        \label{app:xgboost-meta-clf-hyperparameters}
        For the meta-classification experiments~(\Cref{sec:Heterogeneity}), we do not tune hyperparameters but use the fixed configuration in \Cref{tab:xgboost_meta_clf} for all runs, training with a multi-class cross-entropy objective for $300$ boosting rounds.
        \begin{table}[H]
    \renewcommand{\arraystretch}{1.3}
    \centering
    \caption{\textbf{XGBoost hyperparameters for meta-classification}, kept fixed across all meta-classification experiments.}
    \label{tab:xgboost_meta_clf}
    \begin{tabular}{l l}
        \toprule
        \textbf{Hyperparameter} & \textbf{Value} \\
        \midrule
        \texttt{tree\_method}       & \texttt{hist} \\
        \texttt{grow\_policy}       & \texttt{depthwise} \\
        \texttt{learning\_rate}      & 0.1 \\
        \texttt{max\_depth}         & 10 \\
        \texttt{min\_child\_weight} & 0.1 \\
        \texttt{subsample}          & 0.8 \\
        \texttt{colsample\_bytree}  & 0.8 \\
        \texttt{reg\_lambda}        & 1.0 \\
        \texttt{num\_boost\_round}  & 300 \\
        \bottomrule
    \end{tabular}
\end{table}
         
    \paragraph{GIN graph-encoder.}
        \label{app:gin-hyperparameters}
        For the multi-task learning experiments, we use a shared GIN~\citep{xu18a} graph-encoder. Its hyperparameters are held fixed across all dataset pairs, except for the output layer size, which depends on the task head. We optimize with AdamW, using PyTorch defaults except for the learning rate and weight decay, which are set as in \cref{tab:gin_hyperparameters}. Training is stopped early once the validation loss on both datasets has not improved for $100$ epochs.
        \begin{table}[H]
    \renewcommand{\arraystretch}{1.3}
    \centering
    \caption{\textbf{GIN graph-encoder hyperparameters} used in all multi-task experiments. The output dimension of the task head is the only hyperparameter that varies across experiments and is determined by the target dimension of the corresponding dataset.}
    \label{tab:gin_hyperparameters}
    \begin{tabularx}{\textwidth}{X X}
        \toprule
        \textbf{Hyperparameter} & \textbf{Value} \\
        \midrule
        Number of layers                & 4 \\
        Hidden dimension              & 128 \\
        Normalization type                 & Batch \\
        Dropout                         & 0.2 \\
        Pooling                       & sum \\
        Task-head hidden layers       & $[256,\ 128,\ 128]$ \\
        \midrule
        Optimizer                     & AdamW (PyTorch defaults) \\
        Learning rate                 & $10^{-3}$ \\
        Weight decay                  & $10^{-4}$ \\
        Batch size                    & 256 \\
        Early-stopping patience       & 100 \\
        \bottomrule
    \end{tabularx}
\end{table} \clearpage

\section{Additional results}
    In this appendix, we collect extended results, complementing each of the three experiments in the main text.

    \subsection{Expressivity}
\begin{figure}[h]
    \centering
    \includegraphics[width=1.0\textwidth]{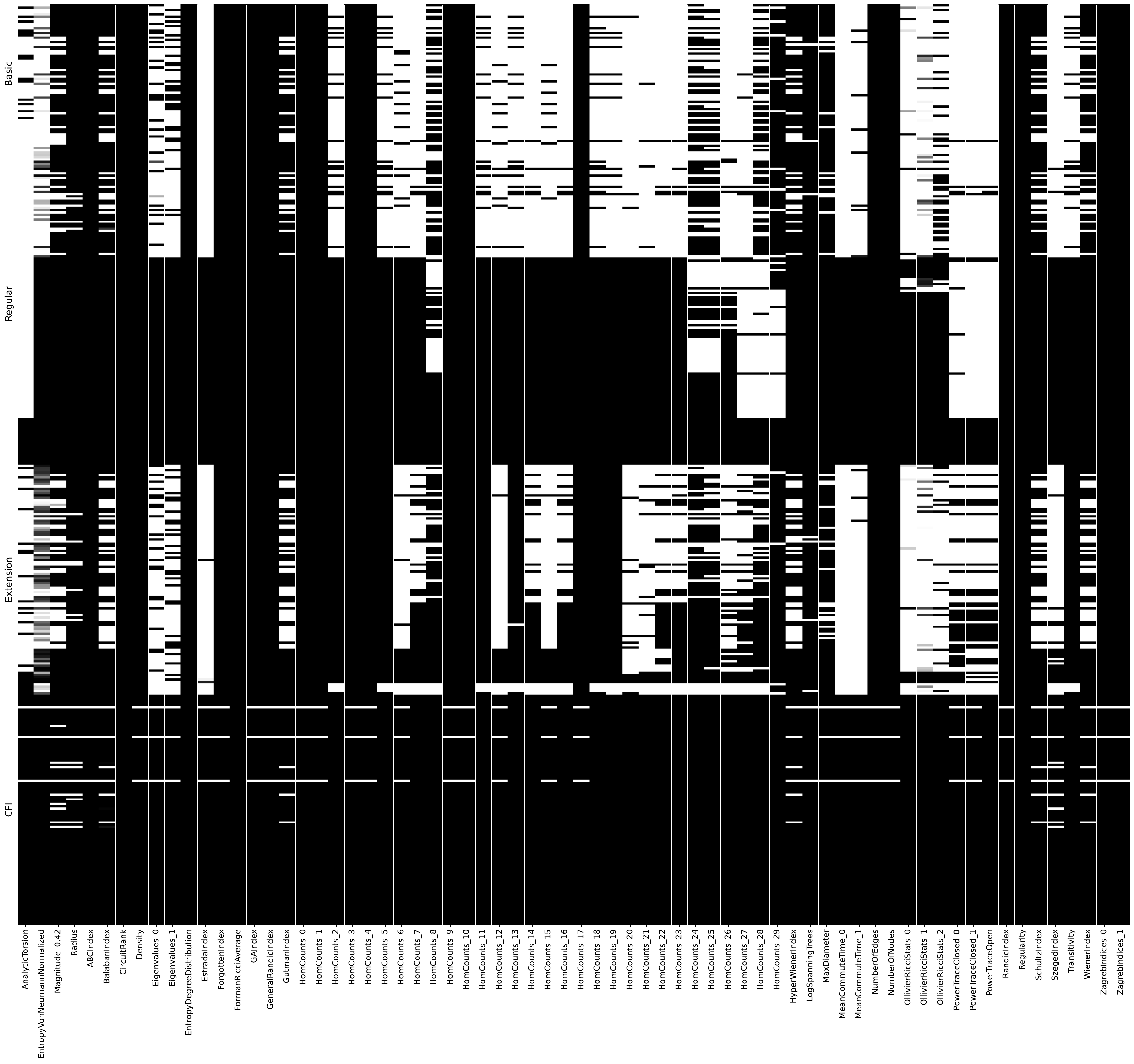}
    \caption{Heatmap that shows BREC graph pairs on the x-axis and invariants on the y-axis. The colour of a cell indicates the difference of an invariant value when computed for both graphs. Brighter cells indicate larger differences. The first four rows form an expressive subset.}
    \label{fig:heatmap_full_invariants}
\end{figure}

\begin{figure}
    \centering
    \includegraphics[width=1.0\textwidth]{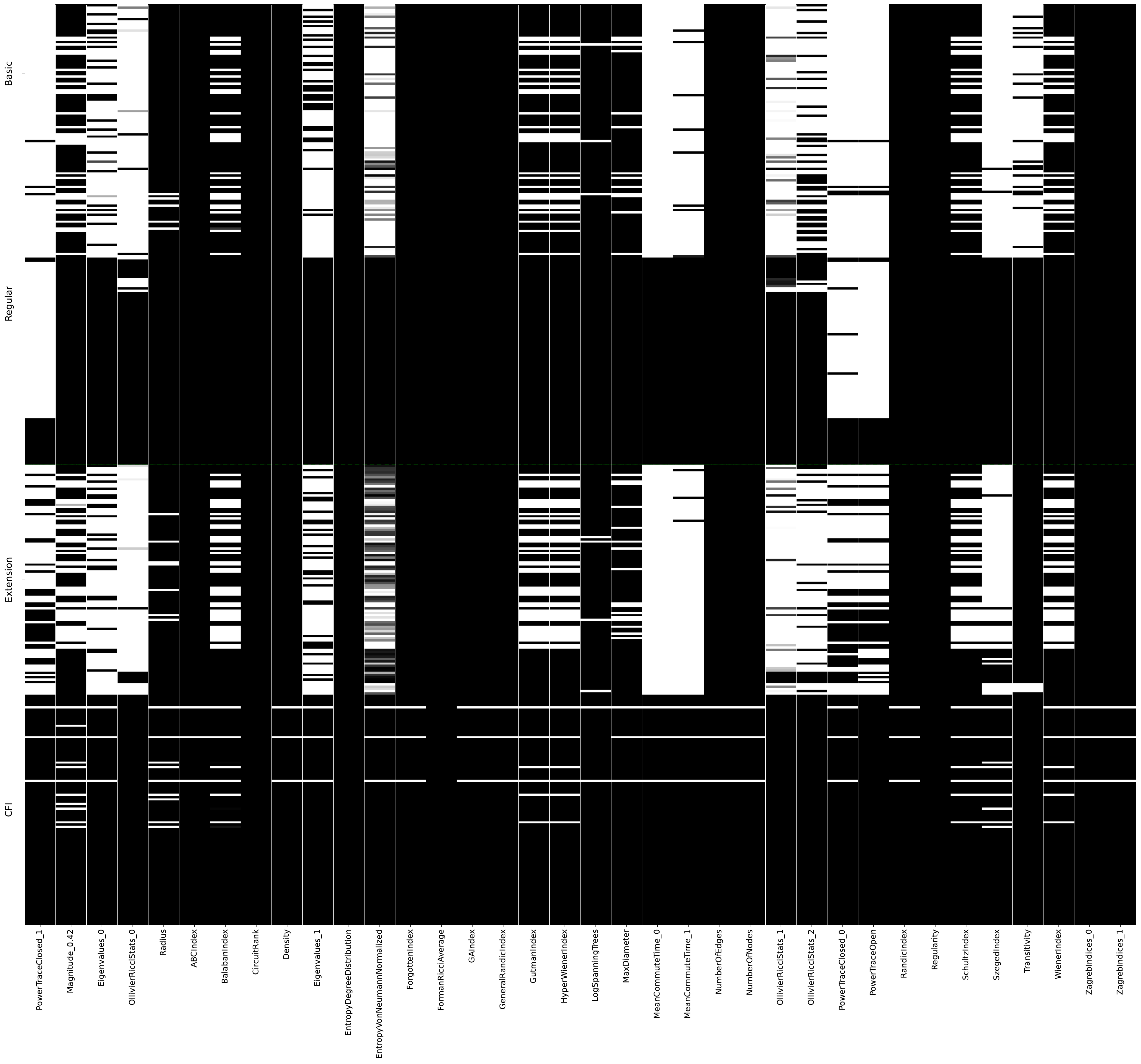}
    \caption{Heatmap that shows BREC graph pairs on the x-axis and invariants on the y-axis. The colour of a cell indicates the difference of an invariant value when computed for both graphs. Brighter cells indicate larger differences. The first four rows form an expressive subset.}
    \label{fig:heatmap_reduced_invariants}
\end{figure}

        Per-invariant expressivity heatmaps for the full and efficient regimes~(\Cref{fig:heatmap_full_invariants,fig:heatmap_reduced_invariants}), complementing~\cref{sec:expressivity}.
                          \clearpage
    
    \subsection{Meta-classification}
            Meta-classification accuracies and confusion matrices, complementing~\cref{sec:Heterogeneity}.

        \paragraph{Matched pairs.}
            \label{app:matched_pairs}
            For the pairwise confounder-free meta-classification on MoleculeNet, \Cref{tab:matched_pairs} lists the number of graphs that can be matched exactly on the confounders, i.e., the number of nodes, the number of edges, and the second degree moment.
Pairs with fewer than $400$ matched graphs~(BACE--QM9, BBBP--QM9, and Lipo--QM9) are omitted from \Cref{tab:confounder_matched_accuracy}.
            \begin{table}[H]
\centering
\caption{\textbf{Exactly matched graphs per dataset pair.} Graphs available per dataset in a cohort matched on the confounders~(number of nodes, number of edges, and second degree moment); the diagonal gives the dataset size.}
\label{tab:matched_pairs}
\small
\setlength{\tabcolsep}{4pt}
\begin{tabular}{l r r r r r r}
\toprule
 & BACE & BBBP & HIV & Lipo & PCBA & QM9 \\
\midrule
BACE & \textit{1{,}513} & 400 & 1{,}317 & 800 & 1{,}427 & 5 \\
BBBP & 400 & \textit{2{,}039} & 1{,}910 & 1{,}284 & 1{,}997 & 209 \\
HIV & 1{,}317 & 1{,}910 & \textit{41{,}127} & 4{,}111 & 37{,}375 & 2{,}406 \\
Lipo & 800 & 1{,}284 & 4{,}111 & \textit{4{,}200} & 4{,}188 & 71 \\
PCBA & 1{,}427 & 1{,}997 & 37{,}375 & 4{,}188 & \textit{437{,}929} & 4{,}018 \\
QM9 & 5 & 209 & 2{,}406 & 71 & 4{,}018 & \textit{130{,}831} \\
\bottomrule
\end{tabular}
\end{table}
 
\begin{figure}
    \centering
    \includegraphics[width=0.41\textwidth]{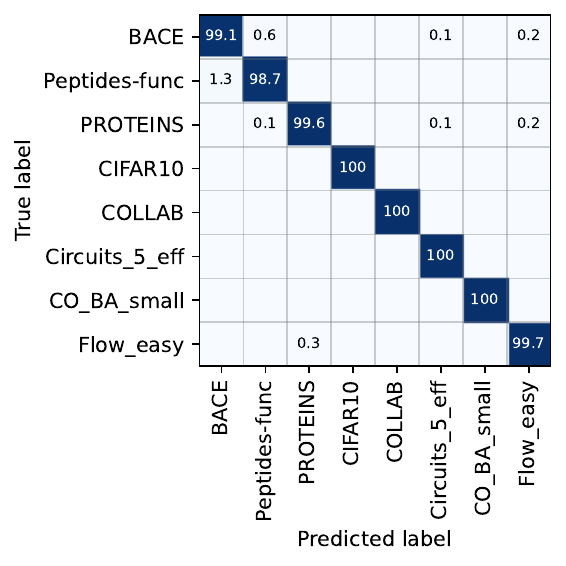}
    \caption{\textbf{Confusion matrix, diverse subset, $\texttt{I}_{\mathrm{eff}}$}. The diverse subset is shown in \Cref{app:diverse_subset} and the invariant set $\texttt{I}_{\mathrm{eff}}$ in \Cref{app:regimes}. The cells show the percentage of each dataset's test graphs assigned to each predicted dataset, aggregated over $5$ seeds; empty cells indicate that no graph was assigned.}
    \label{fig:confusion_diverse}
\end{figure}

        \paragraph{Diverse subset.}
            \label{app:diverse_subset}
            Besides classifying all datasets, we restrict meta-classification to a \emph{diverse subset} comprising one dataset per category of \Cref{tab:datasets_overview}: BACE, Peptides-func, PROTEINS, CIFAR10, COLLAB, Circuits\_5\_eff, CO\_BA\_small, and Flow\_easy.
This selection preserves the structural diversity of our collection in a more compact form; the corresponding confusion matrix is shown in \Cref{fig:confusion_diverse}.

        \begin{table}[H]
    \renewcommand{\arraystretch}{1.3}
    \centering
    \caption{\textbf{Meta-classification results.} Test accuracy, mean $\pm$ standard deviation. Cells show meta-classification results for combinations of datasets and invariant sets with a random sampling of $1{,}000$ graphs per dataset over $5$ seeds. \emph{Diverse} uses one dataset per category~(eight datasets).}
    \label{tab:meta_clf_acc_table}
    \small
    \begin{tabular}{l c c c}
        \toprule
        Datasets & $\texttt{I}_{\mathrm{eff}}$ & $\texttt{S}_{\mathrm{eff}}$ & Confounders \\
        \midrule
        All datasets & $0.612_{\pm 0.006}$ & $0.585_{\pm 0.005}$ & $0.559_{\pm 0.004}$ \\
        Diverse & $0.996_{\pm 0.002}$ & $0.993_{\pm 0.002}$ & $0.989_{\pm 0.003}$ \\
        \bottomrule
    \end{tabular}
\end{table}
         \begin{figure}
    \centering
    \includegraphics[width=0.90\textwidth]{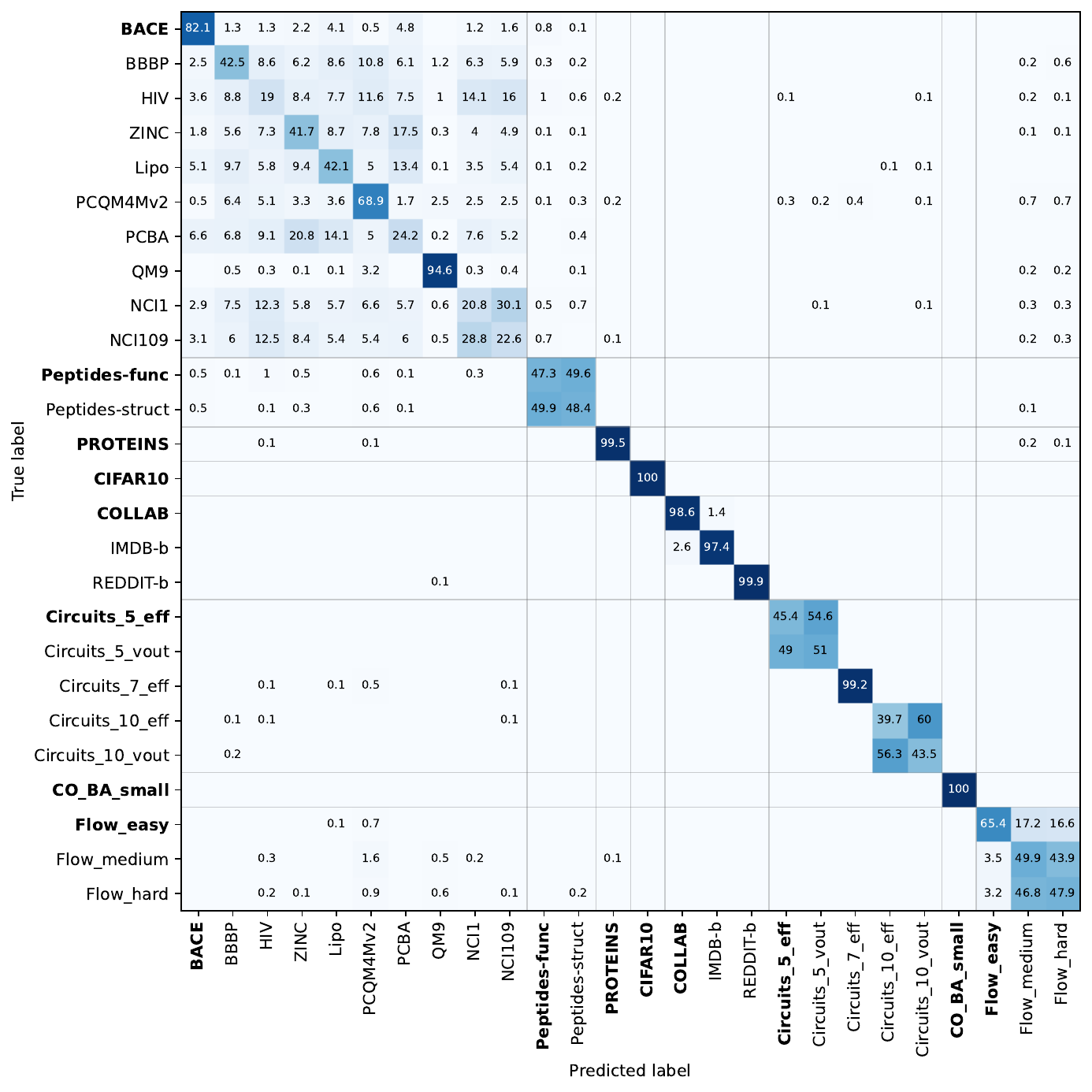}
    \caption{\textbf{Confusion matrix, all datasets, $\texttt{I}_{\mathrm{eff}}$}. The datasets are shown in \Cref{tab:datasets_overview} and the invariant set $\texttt{I}_{\mathrm{eff}}$ in \Cref{app:regimes}. The cells show the percentage of each dataset's test graphs assigned to each predicted dataset, aggregated over $5$ seeds; empty cells indicate that no graph was assigned. Thin lines separate the dataset categories, and datasets of the diverse subset~(\Cref{fig:confusion_diverse}) are shown in bold.}
    \label{fig:confusion_all}
\end{figure}
         \clearpage

    \subsection{Benchmarks}
        Full benchmark tables and domain comparisons referenced in~\cref{sec:relevance_of_structure}. Results reported in the literature for the TU, MoleculeNet, LRGB, and Flow datasets are listed in \Cref{tab:bm_tu,tab:bm_molecules,tab:bm_peptides,tab:bm_flow}.

        \begin{table}[h]
\centering
\caption{XGBoost trained across $26$ datasets and eight input configurations. Each cell shows performance after tuning hyperparameters for $150$ iterations with a Bayesian optimization sampler. Official splits and evaluation procedures were used whenever available. Test metric (\%, mean $\pm$ std) reported over $5$ seeds or $10$ folds for nested cross validation. In each row, the best mean is shown in bold and the second best is underlined, respecting the direction of each metric; tied means share the same marking. For datasets of the efficient regime~(COLLAB, CIFAR10, IMDB-b, REDDIT-b), the invariant sets $\texttt{I}$ and $\texttt{S}$ refer to $\texttt{I}_{\mathrm{eff}}$ and $\texttt{S}_{\mathrm{eff}}$~(\Cref{tab:invariants_subsets}).}
\label{tab:full_table}
\resizebox{\textwidth}{!}{
\begin{tabular}{llllllllll}
\toprule
Metric & Dataset & $\texttt{agg}+\texttt{I}$ & $\texttt{agg}+\texttt{S}$ & \texttt{I} & \texttt{agg} & \texttt{sum} & \texttt{S} & $\texttt{sum}+\texttt{I}$ & $\texttt{sum}+\texttt{S}$ \\
\midrule
\multirow[t]{2}{*}{AP} & PCBA & \textbf{0.1465 $\pm$ 0.0016} & \underline{0.1459 $\pm$ 0.0018} & 0.0704 $\pm$ 0.0003 & 0.1452 $\pm$ 0.0005 & 0.1102 $\pm$ 0.0007 & 0.0554 $\pm$ 0.0009 & 0.1226 $\pm$ 0.0003 & 0.1140 $\pm$ 0.0018 \\
 & Peptides-func & \textbf{0.6874 $\pm$ 0.0022} & 0.6822 $\pm$ 0.0010 & 0.5836 $\pm$ 0.0014 & \underline{0.6838 $\pm$ 0.0043} & 0.6637 $\pm$ 0.0019 & 0.5120 $\pm$ 0.0025 & 0.6765 $\pm$ 0.0021 & 0.6678 $\pm$ 0.0015 \\
\midrule
\multirow[t]{8}{*}{MAE} & PCQM4Mv2 & \underline{0.1520 $\pm$ 0.0000} & 0.1587 $\pm$ 0.0002 & 0.7397 $\pm$ 0.0002 & \textbf{0.1516 $\pm$ 0.0002} & 0.2591 $\pm$ 0.0001 & 0.7641 $\pm$ 0.0003 & 0.2099 $\pm$ 0.0002 & 0.2116 $\pm$ 0.0002 \\
 & QM9 & \textbf{4.0590 $\pm$ 0.0283} & \underline{4.4554 $\pm$ 0.0156} & 90.8535 $\pm$ 0.0757 & 6.1227 $\pm$ 0.0170 & 9.0812 $\pm$ 0.0151 & 151.1581 $\pm$ 0.0548 & 4.4817 $\pm$ 0.0182 & 4.9699 $\pm$ 0.0187 \\
 & ZINC & \textbf{0.4450 $\pm$ 0.0032} & 0.5162 $\pm$ 0.0013 & 1.3188 $\pm$ 0.0018 & 0.5251 $\pm$ 0.0012 & 0.6177 $\pm$ 0.0021 & 1.3643 $\pm$ 0.0015 & \underline{0.4996 $\pm$ 0.0044} & 0.5700 $\pm$ 0.0021 \\
 & CO\_BA\_small & \textbf{1.5531 $\pm$ 0.0012} & 2.4824 $\pm$ 0.0012 & 1.5542 $\pm$ 0.0007 & 2.4559 $\pm$ 0.0019 & 2.9617 $\pm$ 0.0009 & 2.7281 $\pm$ 0.0010 & \underline{1.5535 $\pm$ 0.0012} & 2.7265 $\pm$ 0.0006 \\
 & Flow\_easy & \underline{1.8328 $\pm$ 0.0062} & 1.8597 $\pm$ 0.0307 & 3.7735 $\pm$ 0.0186 & \textbf{1.7674 $\pm$ 0.0039} & 3.6538 $\pm$ 0.0078 & 3.7452 $\pm$ 0.0050 & 3.6133 $\pm$ 0.0000 & 3.6228 $\pm$ 0.0163 \\
 & Flow\_hard & \underline{5.2603 $\pm$ 0.0305} & 5.3428 $\pm$ 0.0000 & 9.2036 $\pm$ 0.1258 & \textbf{5.2001 $\pm$ 0.0287} & 8.2483 $\pm$ 0.0000 & 10.3463 $\pm$ 0.0495 & 8.9556 $\pm$ 0.1325 & 9.6316 $\pm$ 0.0471 \\
 & Flow\_medium & \underline{5.2954 $\pm$ 0.1131} & 5.3698 $\pm$ 0.0417 & 9.4187 $\pm$ 0.3461 & \textbf{5.1565 $\pm$ 0.0532} & 8.0768 $\pm$ 0.0563 & 10.7584 $\pm$ 0.0374 & 9.2379 $\pm$ 0.2110 & 9.8586 $\pm$ 0.1685 \\
 & Peptides-struct & \textbf{0.2476 $\pm$ 0.0002} & 0.2489 $\pm$ 0.0001 & 0.2576 $\pm$ 0.0003 & 0.2518 $\pm$ 0.0002 & 0.2558 $\pm$ 0.0003 & 0.2627 $\pm$ 0.0001 & \underline{0.2479 $\pm$ 0.0003} & 0.2498 $\pm$ 0.0002 \\
\midrule
MSE & Lipo & \textbf{0.6929 $\pm$ 0.0009} & 0.7251 $\pm$ 0.0031 & 1.0200 $\pm$ 0.0120 & 0.7324 $\pm$ 0.0070 & 0.8118 $\pm$ 0.0021 & 1.0206 $\pm$ 0.0108 & \underline{0.7136 $\pm$ 0.0041} & 0.7514 $\pm$ 0.0055 \\
\midrule
\multirow[t]{5}{*}{RSE} & Circuits\_10\_eff & \underline{0.7539 $\pm$ 0.0010} & \textbf{0.7486 $\pm$ 0.0003} & 0.9474 $\pm$ 0.0031 & 0.7627 $\pm$ 0.0000 & 0.8725 $\pm$ 0.0021 & 0.9505 $\pm$ 0.0005 & 0.8511 $\pm$ 0.0004 & 0.8484 $\pm$ 0.0030 \\
 & Circuits\_10\_vout & 0.8456 $\pm$ 0.0030 & \textbf{0.8371 $\pm$ 0.0011} & 0.9667 $\pm$ 0.0003 & \underline{0.8455 $\pm$ 0.0024} & 0.9872 $\pm$ 0.0041 & 0.9738 $\pm$ 0.0009 & 0.9442 $\pm$ 0.0011 & 0.9492 $\pm$ 0.0005 \\
 & Circuits\_5\_eff & \textbf{0.1965 $\pm$ 0.0001} & \textbf{0.1965 $\pm$ 0.0000} & 0.9267 $\pm$ 0.0000 & \underline{0.1986 $\pm$ 0.0000} & 0.7394 $\pm$ 0.0001 & 0.9267 $\pm$ 0.0000 & 0.6892 $\pm$ 0.0000 & 0.6892 $\pm$ 0.0001 \\
 & Circuits\_5\_vout & \textbf{0.2340 $\pm$ 0.0000} & \textbf{0.2340 $\pm$ 0.0000} & 0.9286 $\pm$ 0.0000 & \underline{0.2356 $\pm$ 0.0000} & 0.7495 $\pm$ 0.0000 & 0.9286 $\pm$ 0.0000 & 0.7028 $\pm$ 0.0000 & 0.7029 $\pm$ 0.0000 \\
 & Circuits\_7\_eff & 0.8811 $\pm$ 0.0094 & \underline{0.8696 $\pm$ 0.0054} & 1.0157 $\pm$ 0.0041 & \textbf{0.8690 $\pm$ 0.0023} & 0.9428 $\pm$ 0.0022 & 1.0160 $\pm$ 0.0086 & 0.9600 $\pm$ 0.0058 & 0.9538 $\pm$ 0.0004 \\
\midrule
\multirow[t]{7}{*}{Accuracy} & CIFAR10 & \underline{0.3009 $\pm$ 0.0021} & 0.2873 $\pm$ 0.0010 & 0.1842 $\pm$ 0.0018 & 0.2854 $\pm$ 0.0010 & 0.2845 $\pm$ 0.0014 & 0.1524 $\pm$ 0.0026 & \textbf{0.3022 $\pm$ 0.0019} & 0.2934 $\pm$ 0.0018 \\
 & COLLAB & \underline{0.7992 $\pm$ 0.0122} & 0.7703 $\pm$ 0.0107 & 0.7977 $\pm$ 0.0128 & 0.7474 $\pm$ 0.0113 & 0.5866 $\pm$ 0.0164 & 0.7493 $\pm$ 0.0156 & \textbf{0.8011 $\pm$ 0.0124} & 0.7571 $\pm$ 0.0172 \\
 & IMDB-b & \underline{0.7109 $\pm$ 0.0349} & 0.7020 $\pm$ 0.0275 & 0.7030 $\pm$ 0.0352 & 0.6883 $\pm$ 0.0366 & 0.5610 $\pm$ 0.0483 & 0.6961 $\pm$ 0.0376 & \textbf{0.7150 $\pm$ 0.0394} & 0.7075 $\pm$ 0.0406 \\
 & NCI1 & \textbf{0.7996 $\pm$ 0.0237} & \underline{0.7924 $\pm$ 0.0231} & 0.7530 $\pm$ 0.0274 & 0.7852 $\pm$ 0.0221 & 0.6968 $\pm$ 0.0249 & 0.7085 $\pm$ 0.0219 & 0.7869 $\pm$ 0.0268 & 0.7632 $\pm$ 0.0166 \\
 & NCI109 & \textbf{0.7967 $\pm$ 0.0147} & \underline{0.7857 $\pm$ 0.0098} & 0.7400 $\pm$ 0.0146 & 0.7740 $\pm$ 0.0128 & 0.7074 $\pm$ 0.0132 & 0.6935 $\pm$ 0.0207 & 0.7765 $\pm$ 0.0152 & 0.7647 $\pm$ 0.0114 \\
 & PROTEINS & \underline{0.7728 $\pm$ 0.0389} & 0.7676 $\pm$ 0.0363 & 0.7415 $\pm$ 0.0361 & 0.7665 $\pm$ 0.0296 & 0.7661 $\pm$ 0.0412 & 0.7357 $\pm$ 0.0403 & 0.7714 $\pm$ 0.0399 & \textbf{0.7753 $\pm$ 0.0343} \\
 & REDDIT-b & \textbf{0.9049 $\pm$ 0.0198} & 0.9015 $\pm$ 0.0162 & \underline{0.9031 $\pm$ 0.0209} & 0.8812 $\pm$ 0.0201 & 0.7775 $\pm$ 0.0190 & 0.8567 $\pm$ 0.0223 & \textbf{0.9049 $\pm$ 0.0174} & 0.8843 $\pm$ 0.0192 \\
\midrule
\multirow[t]{3}{*}{ROC-AUC} & BACE & 0.8239 $\pm$ 0.0050 & \underline{0.8363 $\pm$ 0.0048} & \textbf{0.8364 $\pm$ 0.0047} & 0.8278 $\pm$ 0.0091 & 0.8263 $\pm$ 0.0084 & 0.8266 $\pm$ 0.0054 & 0.8264 $\pm$ 0.0041 & 0.8330 $\pm$ 0.0090 \\
 & BBBP & \underline{0.7408 $\pm$ 0.0000} & \textbf{0.7439 $\pm$ 0.0129} & 0.6562 $\pm$ 0.0048 & 0.7404 $\pm$ 0.0027 & 0.7181 $\pm$ 0.0042 & 0.6770 $\pm$ 0.0035 & 0.7329 $\pm$ 0.0000 & 0.7191 $\pm$ 0.0055 \\
 & HIV & \textbf{0.7695 $\pm$ 0.0032} & 0.7626 $\pm$ 0.0032 & 0.7147 $\pm$ 0.0057 & 0.7438 $\pm$ 0.0087 & 0.7402 $\pm$ 0.0097 & 0.7366 $\pm$ 0.0062 & 0.7648 $\pm$ 0.0000 & \underline{0.7661 $\pm$ 0.0053} \\
\bottomrule
\end{tabular}
}
\end{table}         \begin{table}[h]
\caption{XGBoost trained across targets on QM9 and eight input configurations. Each cell shows performance after tuning hyperparameters for $150$ iterations with a bayesian optimization sampler. For each target, a separate model was fit. Test mean absolute error (mean ± std) reported over $5$ seeds. In each row, the lowest mean is shown in bold and the second lowest is underlined.}

\label{tab:qm9}
\resizebox{\textwidth}{!}{
\begin{tabular}{lllllllll}
\toprule
Target & $\texttt{agg}+\texttt{I}$ & $\texttt{agg}+\texttt{S}$ & \texttt{I} & \texttt{agg} & \texttt{sum} & \texttt{S} & $\texttt{sum}+\texttt{I}$ & $\texttt{sum}+\texttt{S}$ \\
\midrule
$\mu ,D$ & \textbf{0.5846 $\pm$ 0.0012} & \underline{0.5961 $\pm$ 0.0012} & 0.7940 $\pm$ 0.0001 & 0.6047 $\pm$ 0.0009 & 0.7634 $\pm$ 0.0006 & 0.8919 $\pm$ 0.0003 & 0.6509 $\pm$ 0.0002 & 0.6948 $\pm$ 0.0000 \\
$\alpha ,a_0^3$ & \textbf{0.4803 $\pm$ 0.0007} & \underline{0.5405 $\pm$ 0.0009} & 1.4409 $\pm$ 0.0011 & 0.6569 $\pm$ 0.0021 & 1.0894 $\pm$ 0.0001 & 2.1792 $\pm$ 0.0011 & 0.5895 $\pm$ 0.0010 & 0.7512 $\pm$ 0.0007 \\
$\epsilon_{HOMO}$, eV & \textbf{0.1200 $\pm$ 0.0001} & \underline{0.1319 $\pm$ 0.0004} & 0.2006 $\pm$ 0.0001 & 0.1413 $\pm$ 0.0004 & 0.2186 $\pm$ 0.0001 & 0.2636 $\pm$ 0.0001 & 0.1473 $\pm$ 0.0003 & 0.1808 $\pm$ 0.0001 \\
$\epsilon_{LUMO}$, eV & \textbf{0.1456 $\pm$ 0.0004} & \underline{0.1501 $\pm$ 0.0002} & 0.3146 $\pm$ 0.0007 & 0.1586 $\pm$ 0.0004 & 0.2895 $\pm$ 0.0001 & 0.4563 $\pm$ 0.0000 & 0.1895 $\pm$ 0.0001 & 0.2257 $\pm$ 0.0004 \\
$\Delta_\epsilon$, eV & \textbf{0.1798 $\pm$ 0.0002} & \underline{0.1946 $\pm$ 0.0005} & 0.3146 $\pm$ 0.0003 & 0.2085 $\pm$ 0.0006 & 0.3626 $\pm$ 0.0002 & 0.4806 $\pm$ 0.0005 & 0.2317 $\pm$ 0.0004 & 0.2856 $\pm$ 0.0002 \\
$\langle R^2 \rangle ,a_0^2$ & \textbf{35.2234 $\pm$ 0.1031} & \underline{40.0148 $\pm$ 0.0000} & 66.4269 $\pm$ 0.0501 & 60.6087 $\pm$ 0.0480 & 102.1372 $\pm$ 0.0090 & 78.9205 $\pm$ 0.0689 & 42.8080 $\pm$ 0.0218 & 52.5202 $\pm$ 0.0714 \\
ZPVE, eV & \textbf{0.0065 $\pm$ 0.0000} & \underline{0.0071 $\pm$ 0.0000} & 0.0315 $\pm$ 0.0000 & 0.0110 $\pm$ 0.0001 & 0.0225 $\pm$ 0.0000 & 0.0404 $\pm$ 0.0001 & 0.0079 $\pm$ 0.0000 & 0.0100 $\pm$ 0.0000 \\
$U_0$, eV & 2.8585 $\pm$ 0.0999 & 2.6446 $\pm$ 0.1476 & 255.2492 $\pm$ 0.2968 & 2.6059 $\pm$ 0.0917 & \textbf{0.7626 $\pm$ 0.0000} & 431.6066 $\pm$ 0.0000 & 2.1957 $\pm$ 0.1375 & \underline{1.1485 $\pm$ 0.0000} \\
$U$, eV & 2.9647 $\pm$ 0.0000 & 2.8989 $\pm$ 0.2647 & 254.1861 $\pm$ 0.2659 & 2.5184 $\pm$ 0.0518 & \textbf{0.7841 $\pm$ 0.0315} & 432.0574 $\pm$ 0.4709 & 2.2673 $\pm$ 0.0000 & \underline{1.3797 $\pm$ 0.1004} \\
$H$, eV & 2.9476 $\pm$ 0.1286 & 2.9254 $\pm$ 0.2682 & 256.4001 $\pm$ 0.0000 & 2.5142 $\pm$ 0.1329 & \textbf{0.8186 $\pm$ 0.0935} & 432.9514 $\pm$ 0.5786 & 2.2147 $\pm$ 0.1574 & \underline{1.0580 $\pm$ 0.1373} \\
$G$, eV & 3.0204 $\pm$ 0.2566 & 3.1680 $\pm$ 0.0816 & 254.3722 $\pm$ 0.3803 & 3.1520 $\pm$ 0.0000 & \textbf{0.9200 $\pm$ 0.1166} & 433.3830 $\pm$ 0.2723 & 2.2124 $\pm$ 0.0750 & \underline{1.0408 $\pm$ 0.0917} \\
$c_v, \frac{cal}{molK}$ & \textbf{0.1762 $\pm$ 0.0004} & \underline{0.1930 $\pm$ 0.0004} & 0.5107 $\pm$ 0.0007 & 0.2925 $\pm$ 0.0004 & 0.8059 $\pm$ 0.0002 & 0.6658 $\pm$ 0.0007 & 0.2659 $\pm$ 0.0004 & 0.3438 $\pm$ 0.0008 \\
\bottomrule
\end{tabular}
}
\end{table}         \clearpage

        \begin{table}[h]
\centering
\caption{$95\%$ percentile bootstrap confidence intervals~($10{,}000$ resamples) for the results in \Cref{tab:full_table}. Each cell shows the mean test metric with its confidence interval in brackets, computed over $5$ seeds or $10$ folds for nested cross validation. Note that cross-validation folds share training data and are therefore not independent. Bootstrap bias is \emph{negligible} in all cells~(max $|\mathrm{bias}|/\mathrm{SE} < 0.03$), where $\mathrm{SE}$ denotes the bootstrap standard error. Intervals with equal bounds are narrower than the displayed precision.}
\label{tab:full_table_bootstrap}
\resizebox{\textwidth}{!}{
\begin{tabular}{llllllllll}
\toprule
Metric & Dataset & $\texttt{agg}+\texttt{I}$ & $\texttt{agg}+\texttt{S}$ & \texttt{I} & \texttt{agg} & \texttt{sum} & \texttt{S} & $\texttt{sum}+\texttt{I}$ & $\texttt{sum}+\texttt{S}$ \\
\midrule
\multirow[t]{2}{*}{AP} & PCBA & 0.147\,{\scriptsize[0.145, 0.148]} & 0.146\,{\scriptsize[0.144, 0.147]} & 0.070\,{\scriptsize[0.070, 0.071]} & 0.145\,{\scriptsize[0.145, 0.146]} & 0.110\,{\scriptsize[0.110, 0.111]} & 0.055\,{\scriptsize[0.055, 0.056]} & 0.123\,{\scriptsize[0.122, 0.123]} & 0.114\,{\scriptsize[0.113, 0.115]} \\
 & Peptides-func & 0.687\,{\scriptsize[0.686, 0.689]} & 0.682\,{\scriptsize[0.681, 0.683]} & 0.584\,{\scriptsize[0.582, 0.585]} & 0.684\,{\scriptsize[0.681, 0.687]} & 0.664\,{\scriptsize[0.662, 0.665]} & 0.512\,{\scriptsize[0.510, 0.514]} & 0.676\,{\scriptsize[0.675, 0.678]} & 0.668\,{\scriptsize[0.667, 0.669]} \\
\midrule
\multirow[t]{8}{*}{MAE} & PCQM4Mv2 & 0.152\,{\scriptsize[0.152, 0.152]} & 0.159\,{\scriptsize[0.159, 0.159]} & 0.740\,{\scriptsize[0.740, 0.740]} & 0.152\,{\scriptsize[0.151, 0.152]} & 0.259\,{\scriptsize[0.259, 0.259]} & 0.764\,{\scriptsize[0.764, 0.764]} & 0.210\,{\scriptsize[0.210, 0.210]} & 0.212\,{\scriptsize[0.211, 0.212]} \\
 & QM9 & 4.059\,{\scriptsize[4.038, 4.081]} & 4.455\,{\scriptsize[4.441, 4.466]} & 90.853\,{\scriptsize[90.799, 90.914]} & 6.123\,{\scriptsize[6.110, 6.135]} & 9.081\,{\scriptsize[9.068, 9.091]} & 151.158\,{\scriptsize[151.111, 151.198]} & 4.482\,{\scriptsize[4.466, 4.494]} & 4.970\,{\scriptsize[4.954, 4.984]} \\
 & ZINC & 0.445\,{\scriptsize[0.442, 0.447]} & 0.516\,{\scriptsize[0.515, 0.517]} & 1.319\,{\scriptsize[1.317, 1.320]} & 0.525\,{\scriptsize[0.524, 0.526]} & 0.618\,{\scriptsize[0.616, 0.620]} & 1.364\,{\scriptsize[1.363, 1.365]} & 0.500\,{\scriptsize[0.496, 0.504]} & 0.570\,{\scriptsize[0.568, 0.572]} \\
 & CO\_BA\_small & 1.553\,{\scriptsize[1.552, 1.554]} & 2.482\,{\scriptsize[2.481, 2.484]} & 1.554\,{\scriptsize[1.554, 1.555]} & 2.456\,{\scriptsize[2.454, 2.458]} & 2.962\,{\scriptsize[2.961, 2.962]} & 2.728\,{\scriptsize[2.727, 2.729]} & 1.553\,{\scriptsize[1.552, 1.554]} & 2.726\,{\scriptsize[2.726, 2.727]} \\
 & Flow\_easy & 1.833\,{\scriptsize[1.829, 1.839]} & 1.860\,{\scriptsize[1.837, 1.887]} & 3.773\,{\scriptsize[3.758, 3.789]} & 1.767\,{\scriptsize[1.764, 1.770]} & 3.654\,{\scriptsize[3.648, 3.660]} & 3.745\,{\scriptsize[3.741, 3.749]} & 3.613\,{\scriptsize[3.613, 3.613]} & 3.623\,{\scriptsize[3.609, 3.635]} \\
 & Flow\_hard & 5.260\,{\scriptsize[5.237, 5.288]} & 5.343\,{\scriptsize[5.343, 5.343]} & 9.204\,{\scriptsize[9.107, 9.300]} & 5.200\,{\scriptsize[5.179, 5.225]} & 8.248\,{\scriptsize[8.248, 8.248]} & 10.346\,{\scriptsize[10.304, 10.384]} & 8.956\,{\scriptsize[8.843, 9.062]} & 9.632\,{\scriptsize[9.593, 9.670]} \\
 & Flow\_medium & 5.295\,{\scriptsize[5.199, 5.383]} & 5.370\,{\scriptsize[5.337, 5.406]} & 9.419\,{\scriptsize[9.115, 9.717]} & 5.157\,{\scriptsize[5.104, 5.190]} & 8.077\,{\scriptsize[8.035, 8.123]} & 10.758\,{\scriptsize[10.730, 10.790]} & 9.238\,{\scriptsize[9.040, 9.379]} & 9.859\,{\scriptsize[9.732, 10.005]} \\
 & Peptides-struct & 0.248\,{\scriptsize[0.247, 0.248]} & 0.249\,{\scriptsize[0.249, 0.249]} & 0.258\,{\scriptsize[0.257, 0.258]} & 0.252\,{\scriptsize[0.252, 0.252]} & 0.256\,{\scriptsize[0.256, 0.256]} & 0.263\,{\scriptsize[0.263, 0.263]} & 0.248\,{\scriptsize[0.248, 0.248]} & 0.250\,{\scriptsize[0.250, 0.250]} \\
\midrule
MSE & Lipo & 0.693\,{\scriptsize[0.692, 0.693]} & 0.725\,{\scriptsize[0.722, 0.727]} & 1.020\,{\scriptsize[1.010, 1.030]} & 0.732\,{\scriptsize[0.727, 0.738]} & 0.812\,{\scriptsize[0.810, 0.814]} & 1.021\,{\scriptsize[1.010, 1.027]} & 0.714\,{\scriptsize[0.711, 0.717]} & 0.751\,{\scriptsize[0.747, 0.755]} \\
\midrule
\multirow[t]{5}{*}{RSE} & Circuits\_10\_eff & 0.754\,{\scriptsize[0.753, 0.755]} & 0.749\,{\scriptsize[0.748, 0.749]} & 0.947\,{\scriptsize[0.945, 0.950]} & 0.763\,{\scriptsize[0.763, 0.763]} & 0.873\,{\scriptsize[0.871, 0.874]} & 0.951\,{\scriptsize[0.950, 0.951]} & 0.851\,{\scriptsize[0.851, 0.851]} & 0.848\,{\scriptsize[0.845, 0.850]} \\
 & Circuits\_10\_vout & 0.846\,{\scriptsize[0.843, 0.848]} & 0.837\,{\scriptsize[0.836, 0.838]} & 0.967\,{\scriptsize[0.966, 0.967]} & 0.846\,{\scriptsize[0.843, 0.847]} & 0.987\,{\scriptsize[0.984, 0.991]} & 0.974\,{\scriptsize[0.973, 0.975]} & 0.944\,{\scriptsize[0.943, 0.945]} & 0.949\,{\scriptsize[0.949, 0.950]} \\
 & Circuits\_5\_eff & 0.196\,{\scriptsize[0.196, 0.197]} & 0.196\,{\scriptsize[0.196, 0.196]} & 0.927\,{\scriptsize[0.927, 0.927]} & 0.199\,{\scriptsize[0.199, 0.199]} & 0.739\,{\scriptsize[0.739, 0.739]} & 0.927\,{\scriptsize[0.927, 0.927]} & 0.689\,{\scriptsize[0.689, 0.689]} & 0.689\,{\scriptsize[0.689, 0.689]} \\
 & Circuits\_5\_vout & 0.234\,{\scriptsize[0.234, 0.234]} & 0.234\,{\scriptsize[0.234, 0.234]} & 0.929\,{\scriptsize[0.929, 0.929]} & 0.236\,{\scriptsize[0.236, 0.236]} & 0.750\,{\scriptsize[0.749, 0.750]} & 0.929\,{\scriptsize[0.929, 0.929]} & 0.703\,{\scriptsize[0.703, 0.703]} & 0.703\,{\scriptsize[0.703, 0.703]} \\
 & Circuits\_7\_eff & 0.881\,{\scriptsize[0.873, 0.889]} & 0.870\,{\scriptsize[0.865, 0.874]} & 1.016\,{\scriptsize[1.012, 1.019]} & 0.869\,{\scriptsize[0.867, 0.871]} & 0.943\,{\scriptsize[0.941, 0.944]} & 1.016\,{\scriptsize[1.009, 1.022]} & 0.960\,{\scriptsize[0.956, 0.965]} & 0.954\,{\scriptsize[0.953, 0.954]} \\
\midrule
\multirow[t]{7}{*}{Accuracy} & CIFAR10 & 0.301\,{\scriptsize[0.299, 0.302]} & 0.287\,{\scriptsize[0.287, 0.288]} & 0.184\,{\scriptsize[0.184, 0.185]} & 0.285\,{\scriptsize[0.285, 0.286]} & 0.285\,{\scriptsize[0.284, 0.285]} & 0.152\,{\scriptsize[0.152, 0.153]} & 0.302\,{\scriptsize[0.301, 0.304]} & 0.293\,{\scriptsize[0.293, 0.294]} \\
 & COLLAB & 0.799\,{\scriptsize[0.792, 0.806]} & 0.770\,{\scriptsize[0.765, 0.775]} & 0.798\,{\scriptsize[0.790, 0.805]} & 0.747\,{\scriptsize[0.742, 0.753]} & 0.587\,{\scriptsize[0.577, 0.595]} & 0.749\,{\scriptsize[0.745, 0.754]} & 0.801\,{\scriptsize[0.795, 0.809]} & 0.757\,{\scriptsize[0.751, 0.763]} \\
 & IMDB-b & 0.711\,{\scriptsize[0.690, 0.730]} & 0.702\,{\scriptsize[0.689, 0.714]} & 0.703\,{\scriptsize[0.682, 0.721]} & 0.688\,{\scriptsize[0.667, 0.707]} & 0.561\,{\scriptsize[0.535, 0.591]} & 0.696\,{\scriptsize[0.676, 0.716]} & 0.715\,{\scriptsize[0.693, 0.738]} & 0.707\,{\scriptsize[0.687, 0.730]} \\
 & NCI1 & 0.800\,{\scriptsize[0.786, 0.814]} & 0.792\,{\scriptsize[0.779, 0.806]} & 0.753\,{\scriptsize[0.736, 0.768]} & 0.785\,{\scriptsize[0.773, 0.798]} & 0.697\,{\scriptsize[0.682, 0.712]} & 0.709\,{\scriptsize[0.696, 0.722]} & 0.787\,{\scriptsize[0.772, 0.803]} & 0.763\,{\scriptsize[0.754, 0.774]} \\
 & NCI109 & 0.797\,{\scriptsize[0.788, 0.805]} & 0.786\,{\scriptsize[0.781, 0.791]} & 0.740\,{\scriptsize[0.732, 0.749]} & 0.774\,{\scriptsize[0.767, 0.780]} & 0.707\,{\scriptsize[0.700, 0.715]} & 0.693\,{\scriptsize[0.681, 0.705]} & 0.776\,{\scriptsize[0.768, 0.785]} & 0.765\,{\scriptsize[0.759, 0.771]} \\
 & PROTEINS & 0.773\,{\scriptsize[0.750, 0.796]} & 0.768\,{\scriptsize[0.748, 0.789]} & 0.741\,{\scriptsize[0.720, 0.763]} & 0.766\,{\scriptsize[0.750, 0.783]} & 0.766\,{\scriptsize[0.744, 0.790]} & 0.736\,{\scriptsize[0.713, 0.760]} & 0.771\,{\scriptsize[0.748, 0.795]} & 0.775\,{\scriptsize[0.756, 0.795]} \\
 & REDDIT-b & 0.905\,{\scriptsize[0.894, 0.917]} & 0.902\,{\scriptsize[0.893, 0.911]} & 0.903\,{\scriptsize[0.891, 0.915]} & 0.881\,{\scriptsize[0.869, 0.892]} & 0.778\,{\scriptsize[0.766, 0.789]} & 0.857\,{\scriptsize[0.846, 0.868]} & 0.905\,{\scriptsize[0.895, 0.915]} & 0.884\,{\scriptsize[0.872, 0.895]} \\
\midrule
\multirow[t]{3}{*}{ROC-AUC} & BACE & 0.824\,{\scriptsize[0.820, 0.828]} & 0.836\,{\scriptsize[0.833, 0.840]} & 0.836\,{\scriptsize[0.833, 0.840]} & 0.828\,{\scriptsize[0.821, 0.836]} & 0.826\,{\scriptsize[0.820, 0.833]} & 0.827\,{\scriptsize[0.822, 0.831]} & 0.826\,{\scriptsize[0.823, 0.830]} & 0.833\,{\scriptsize[0.826, 0.841]} \\
 & BBBP & 0.741\,{\scriptsize[0.741, 0.741]} & 0.744\,{\scriptsize[0.733, 0.755]} & 0.656\,{\scriptsize[0.653, 0.660]} & 0.740\,{\scriptsize[0.738, 0.743]} & 0.718\,{\scriptsize[0.715, 0.722]} & 0.677\,{\scriptsize[0.674, 0.680]} & 0.733\,{\scriptsize[0.733, 0.733]} & 0.719\,{\scriptsize[0.715, 0.724]} \\
 & HIV & 0.769\,{\scriptsize[0.767, 0.772]} & 0.763\,{\scriptsize[0.760, 0.765]} & 0.715\,{\scriptsize[0.710, 0.719]} & 0.744\,{\scriptsize[0.736, 0.752]} & 0.740\,{\scriptsize[0.731, 0.748]} & 0.737\,{\scriptsize[0.731, 0.742]} & 0.765\,{\scriptsize[0.765, 0.765]} & 0.766\,{\scriptsize[0.761, 0.770]} \\
\bottomrule
\end{tabular}
}
\end{table}
         \begin{table}[h]
\centering
\caption{$95\%$ percentile bootstrap confidence intervals~($10{,}000$ resamples) for the results in \Cref{tab:qm9}. Each cell shows the test mean absolute error with its confidence interval in brackets, computed over $5$ seeds. Bootstrap bias is negligible in all cells~(max $|\mathrm{bias}|/\mathrm{SE} < 0.03$), where $\mathrm{SE}$ denotes the bootstrap standard error. Intervals with equal bounds are narrower than the displayed precision.}
\label{tab:qm9_bootstrap}
\resizebox{\textwidth}{!}{
\begin{tabular}{lllllllll}
\toprule
Target & $\texttt{agg}+\texttt{I}$ & $\texttt{agg}+\texttt{S}$ & \texttt{I} & \texttt{agg} & \texttt{sum} & \texttt{S} & $\texttt{sum}+\texttt{I}$ & $\texttt{sum}+\texttt{S}$ \\
\midrule
$\mu ,D$ & 0.585\,{\scriptsize[0.584, 0.586]} & 0.596\,{\scriptsize[0.595, 0.597]} & 0.794\,{\scriptsize[0.794, 0.794]} & 0.605\,{\scriptsize[0.604, 0.606]} & 0.763\,{\scriptsize[0.763, 0.764]} & 0.892\,{\scriptsize[0.892, 0.892]} & 0.651\,{\scriptsize[0.651, 0.651]} & 0.695\,{\scriptsize[0.695, 0.695]} \\
$\alpha ,a_0^3$ & 0.480\,{\scriptsize[0.480, 0.481]} & 0.540\,{\scriptsize[0.540, 0.541]} & 1.441\,{\scriptsize[1.440, 1.442]} & 0.657\,{\scriptsize[0.655, 0.659]} & 1.089\,{\scriptsize[1.089, 1.090]} & 2.179\,{\scriptsize[2.178, 2.180]} & 0.590\,{\scriptsize[0.589, 0.590]} & 0.751\,{\scriptsize[0.751, 0.752]} \\
$\epsilon_{HOMO}$, eV & 0.120\,{\scriptsize[0.120, 0.120]} & 0.132\,{\scriptsize[0.132, 0.132]} & 0.201\,{\scriptsize[0.201, 0.201]} & 0.141\,{\scriptsize[0.141, 0.142]} & 0.219\,{\scriptsize[0.219, 0.219]} & 0.264\,{\scriptsize[0.264, 0.264]} & 0.147\,{\scriptsize[0.147, 0.147]} & 0.181\,{\scriptsize[0.181, 0.181]} \\
$\epsilon_{LUMO}$, eV & 0.146\,{\scriptsize[0.145, 0.146]} & 0.150\,{\scriptsize[0.150, 0.150]} & 0.315\,{\scriptsize[0.314, 0.315]} & 0.159\,{\scriptsize[0.158, 0.159]} & 0.289\,{\scriptsize[0.289, 0.290]} & 0.456\,{\scriptsize[0.456, 0.456]} & 0.190\,{\scriptsize[0.189, 0.190]} & 0.226\,{\scriptsize[0.225, 0.226]} \\
$\Delta_\epsilon$, eV & 0.180\,{\scriptsize[0.180, 0.180]} & 0.195\,{\scriptsize[0.194, 0.195]} & 0.315\,{\scriptsize[0.314, 0.315]} & 0.208\,{\scriptsize[0.208, 0.209]} & 0.363\,{\scriptsize[0.362, 0.363]} & 0.481\,{\scriptsize[0.480, 0.481]} & 0.232\,{\scriptsize[0.231, 0.232]} & 0.286\,{\scriptsize[0.285, 0.286]} \\
$\langle R^2 \rangle ,a_0^2$ & 35.223\,{\scriptsize[35.127, 35.302]} & 40.015\,{\scriptsize[40.015, 40.015]} & 66.427\,{\scriptsize[66.381, 66.462]} & 60.609\,{\scriptsize[60.564, 60.645]} & 102.137\,{\scriptsize[102.129, 102.144]} & 78.920\,{\scriptsize[78.857, 78.964]} & 42.808\,{\scriptsize[42.791, 42.828]} & 52.520\,{\scriptsize[52.459, 52.578]} \\
ZPVE, eV & 0.006\,{\scriptsize[0.006, 0.006]} & 0.007\,{\scriptsize[0.007, 0.007]} & 0.031\,{\scriptsize[0.031, 0.032]} & 0.011\,{\scriptsize[0.011, 0.011]} & 0.022\,{\scriptsize[0.022, 0.022]} & 0.040\,{\scriptsize[0.040, 0.040]} & 0.008\,{\scriptsize[0.008, 0.008]} & 0.010\,{\scriptsize[0.010, 0.010]} \\
$U_0$, eV & 2.858\,{\scriptsize[2.775, 2.937]} & 2.645\,{\scriptsize[2.514, 2.752]} & 255.249\,{\scriptsize[255.004, 255.483]} & 2.606\,{\scriptsize[2.527, 2.674]} & 0.763\,{\scriptsize[0.763, 0.763]} & 431.607\,{\scriptsize[431.607, 431.607]} & 2.196\,{\scriptsize[2.081, 2.310]} & 1.149\,{\scriptsize[1.149, 1.149]} \\
$U$, eV & 2.965\,{\scriptsize[2.965, 2.965]} & 2.899\,{\scriptsize[2.679, 3.119]} & 254.186\,{\scriptsize[253.993, 254.427]} & 2.518\,{\scriptsize[2.470, 2.555]} & 0.784\,{\scriptsize[0.756, 0.809]} & 432.057\,{\scriptsize[431.711, 432.472]} & 2.267\,{\scriptsize[2.267, 2.267]} & 1.380\,{\scriptsize[1.285, 1.442]} \\
$H$, eV & 2.948\,{\scriptsize[2.839, 3.046]} & 2.925\,{\scriptsize[2.681, 3.133]} & 256.400\,{\scriptsize[256.400, 256.400]} & 2.514\,{\scriptsize[2.400, 2.619]} & 0.819\,{\scriptsize[0.742, 0.896]} & 432.951\,{\scriptsize[432.440, 433.400]} & 2.215\,{\scriptsize[2.066, 2.320]} & 1.058\,{\scriptsize[0.949, 1.167]} \\
$G$, eV & 3.020\,{\scriptsize[2.838, 3.259]} & 3.168\,{\scriptsize[3.101, 3.235]} & 254.372\,{\scriptsize[254.074, 254.714]} & 3.152\,{\scriptsize[3.152, 3.152]} & 0.920\,{\scriptsize[0.830, 1.028]} & 433.383\,{\scriptsize[433.134, 433.605]} & 2.212\,{\scriptsize[2.145, 2.274]} & 1.041\,{\scriptsize[0.975, 1.126]} \\
$c_v, \frac{cal}{molK}$ & 0.176\,{\scriptsize[0.176, 0.177]} & 0.193\,{\scriptsize[0.193, 0.193]} & 0.511\,{\scriptsize[0.510, 0.511]} & 0.292\,{\scriptsize[0.292, 0.293]} & 0.806\,{\scriptsize[0.806, 0.806]} & 0.666\,{\scriptsize[0.665, 0.666]} & 0.266\,{\scriptsize[0.266, 0.266]} & 0.344\,{\scriptsize[0.343, 0.344]} \\
\bottomrule
\end{tabular}
}
\end{table}
         \clearpage

        \begin{figure}[tbp]
    \centering
    \includegraphics[width=1.0\textwidth]{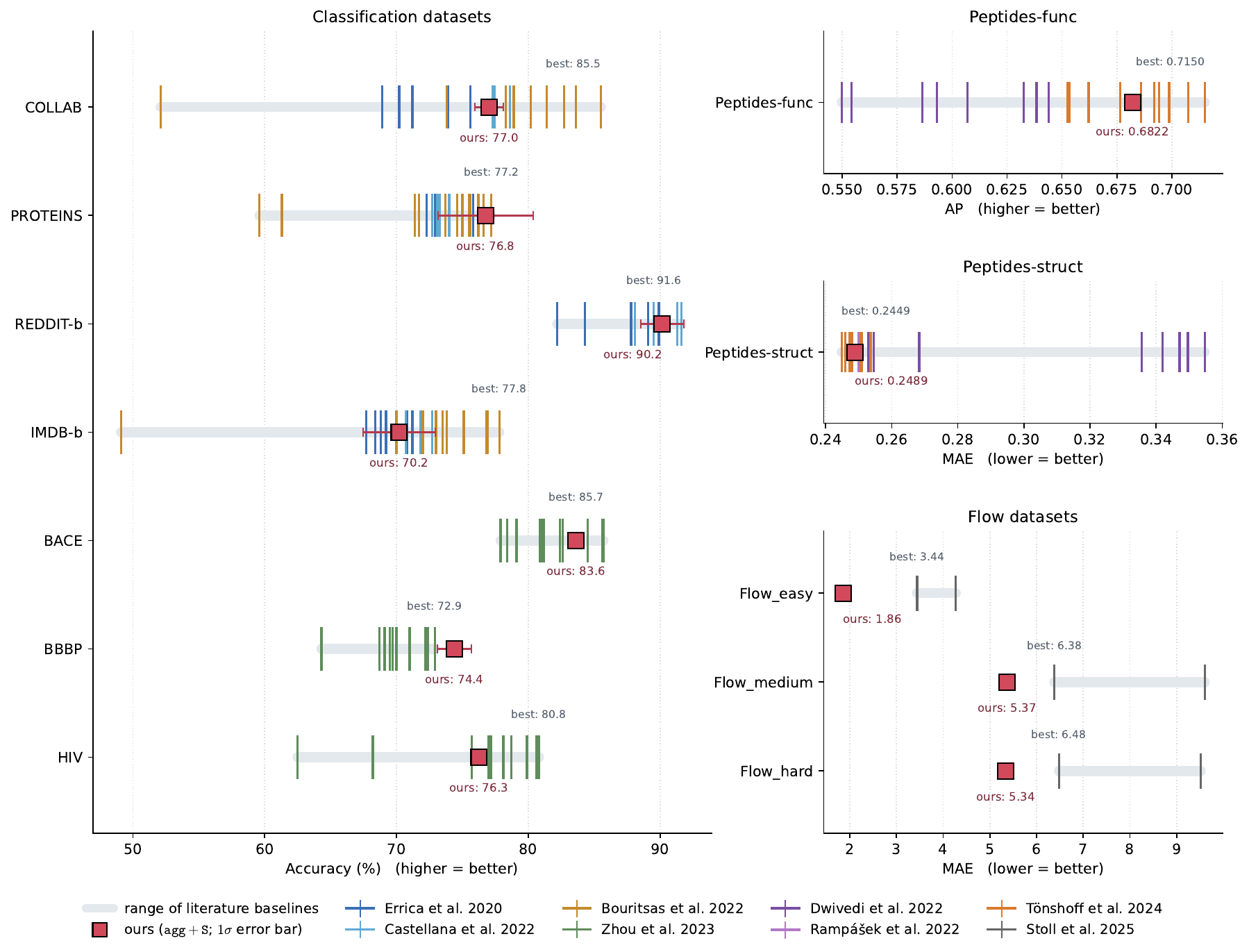}
    \caption{Same comparison as in \Cref{fig:benchmark_comparison}, but for $\texttt{agg}+\texttt{S}$, i.e., using the expressive subset of invariants. Tick colours correspond to source papers of reported performance values. Numeric values are in \Cref{tab:bm_tu} and \Cref{tab:bm_molecules} for classification datasets, in \Cref{tab:bm_peptides} for Peptides and in \Cref{tab:bm_flow} for flow datasets.}
    \label{fig:benchmark_comparison_subset}
\end{figure}
         \begin{table*}[t]
\centering
\small
\setlength{\tabcolsep}{4pt}
\caption{\textbf{TU graph classification.} Test accuracy (\%, mean $\pm$ std). Numbers are taken from the cited papers. We include every model that reports a value on at least one of the four datasets. "OOR" marks Out-Of-Resource entries reported in the source. Bold entries mark the single best reported value in a column.}
\label{tab:bm_tu}
\begin{tabular}{l c c c c}
\toprule
 & \multicolumn{4}{c}{\textit{Accuracy (\%) $\uparrow$}}\\
\cmidrule(lr){2-5}
Model & \textbf{COLLAB} & \textbf{PROTEINS} & \textbf{REDDIT-b} & \textbf{IMDB-b} \\
\midrule
\multicolumn{5}{l}{\citet{Errica2020A}} \\
\addlinespace[2pt]
Baseline (MLP) & 70.2 $\pm$ 1.5 & 75.8 $\pm$ 3.7 & 82.2 $\pm$ 3.0 & 70.8 $\pm$ 5.0 \\
DGCNN & 71.2 $\pm$ 1.9 & 72.9 $\pm$ 3.5 & 87.8 $\pm$ 2.5 & 69.2 $\pm$ 3.0 \\
DiffPool & 68.9 $\pm$ 2.0 & 73.7 $\pm$ 3.5 & 89.1 $\pm$ 1.6 & 68.4 $\pm$ 3.3 \\
ECC & \textit{OOR} & 72.3 $\pm$ 3.4 & \textit{OOR} & 67.7 $\pm$ 2.8 \\
GIN & 75.6 $\pm$ 2.3 & 73.3 $\pm$ 4.0 & 89.9 $\pm$ 1.9 & 71.2 $\pm$ 3.9 \\
GraphSAGE & 73.9 $\pm$ 1.7 & 73.0 $\pm$ 4.5 & 84.3 $\pm$ 1.9 & 68.8 $\pm$ 4.5 \\
\midrule
\multicolumn{5}{l}{\citet{castellana22a}} \\
\addlinespace[2pt]
CGMM & 77.32 $\pm$ 2.2 & 74.0 $\pm$ 3.9 & 88.1 $\pm$ 1.9 & 72.7 $\pm$ 3.6 \\
E-CGMM & 77.45 $\pm$ 2.3 & 73.3 $\pm$ 4.1 & 89.5 $\pm$ 1.3 & 70.7 $\pm$ 3.8 \\
ICGMM$_{\alpha\gamma}$ & --- & 72.7 $\pm$ 3.4 & --- & --- \\
ICGMM$^{f}_{\alpha\gamma}$ & 78.6 $\pm$ 2.8 & 73.3 $\pm$ 2.9 & 91.3 $\pm$ 1.8 & 73.0 $\pm$ 4.3 \\
ICGMM & --- & 73.1 $\pm$ 3.9 & --- & --- \\
ICGMM$^{f}$ & 78.9 $\pm$ 1.7 & 73.2 $\pm$ 3.9 & \textbf{91.6 $\pm$ 2.1} & 71.8 $\pm$ 4.4 \\
\midrule
\multicolumn{5}{l}{\citet{Bouritsas23a}} \\
\addlinespace[2pt]
RWK & --- & 59.6 $\pm$ 0.1 & --- & --- \\
GK (k=3) & --- & 71.4 $\pm$ 0.31 & --- & --- \\
PK & --- & 73.7 $\pm$ 0.7 & --- & --- \\
WL kernel & 78.9 $\pm$ 1.9 & 75.0 $\pm$ 3.1 & --- & 73.8 $\pm$ 3.9 \\
GNTK & 83.6 $\pm$ 1.0 & 75.6 $\pm$ 4.2 & --- & 76.9 $\pm$ 3.6 \\
DCNN & 52.1 $\pm$ 0.7 & 61.3 $\pm$ 1.6 & --- & 49.1 $\pm$ 1.4 \\
DGCNN & 73.8 $\pm$ 0.5 & 75.5 $\pm$ 0.9 & --- & 70.0 $\pm$ 0.9 \\
IGN & 78.3 $\pm$ 2.5 & 76.6 $\pm$ 5.5 & --- & 72.0 $\pm$ 5.5 \\
GIN & 80.2 $\pm$ 1.9 & 76.2 $\pm$ 2.8 & --- & 75.1 $\pm$ 5.1 \\
PPGN & 81.4 $\pm$ 1.4 & 77.2 $\pm$ 4.7 & --- & 73.0 $\pm$ 5.8 \\
Natural GN & --- & 71.7 $\pm$ 1.04 & --- & 73.5 $\pm$ 2.01 \\
GSN-e & \textbf{85.5 $\pm$ 1.2} & 76.6 $\pm$ 5.0 & --- & \textbf{77.8 $\pm$ 3.3} \\
GSN-v & 82.7 $\pm$ 1.5 & 74.59 $\pm$ 5.0 & --- & 76.8 $\pm$ 2.0 \\
\midrule
$\texttt{agg}+\texttt{I}$ (ours) & 79.74 $\pm$ 1.46 & \textbf{77.35 $\pm$ 3.67} & 90.7 $\pm$ 2.02 & 71.4 $\pm$ 3.69 \\
\bottomrule
\end{tabular}
\end{table*}         \begin{table*}[t]
\centering
\small
\caption{\textbf{MoleculeNet.} Test area under receiver operating characteristic curve (\%, mean $\pm$ std evaluated on a scaffold-split). Numbers are taken from the cited paper. We include every model that reports a value on at least one of the three datasets. Bold entries mark the single best reported value in a column.}
\label{tab:bm_molecules}
\setlength{\tabcolsep}{4pt}
\begin{tabular}{l c c c}
\toprule
 & \multicolumn{3}{c}{\textit{AUROC (\%) $\uparrow$ (scaffold split)}}\\
\cmidrule(lr){2-4}
Model & \textbf{BACE} & \textbf{BBBP} & \textbf{HIV} \\
\midrule
\multicolumn{4}{l}{\citet{zhou2023a}} \\
\addlinespace[2pt]
D-MPNN & 80.9 $\pm$ 0.6 & 71.0 $\pm$ 0.3 & 77.1 $\pm$ 0.5 \\
AttentiveFP & 78.4 $\pm$ 0.0 & 64.3 $\pm$ 1.8 & 75.7 $\pm$ 1.4 \\
N-Gram$_{\mathrm{RF}}$ & 77.9 $\pm$ 1.5 & 69.7 $\pm$ 0.6 & 77.2 $\pm$ 0.1 \\
N-Gram$_{\mathrm{XGB}}$ & 79.1 $\pm$ 1.3 & 69.1 $\pm$ 0.8 & 78.7 $\pm$ 0.4 \\
PretrainGNN & 84.5 $\pm$ 0.7 & 68.7 $\pm$ 1.3 & 79.9 $\pm$ 0.7 \\
GROVER$_{\mathrm{base}}$ & 82.6 $\pm$ 0.7 & 70.0 $\pm$ 0.1 & 62.5 $\pm$ 0.9 \\
GROVER$_{\mathrm{large}}$ & 81.0 $\pm$ 1.4 & 69.5 $\pm$ 0.1 & 68.2 $\pm$ 1.1 \\
GraphMVP & 81.2 $\pm$ 0.9 & 72.4 $\pm$ 1.6 & 77.0 $\pm$ 1.2 \\
MolCLR & 82.4 $\pm$ 0.9 & 72.2 $\pm$ 2.1 & 78.1 $\pm$ 0.5 \\
GEM & 85.6 $\pm$ 1.1 & 72.4 $\pm$ 0.4 & 80.6 $\pm$ 0.9 \\
Uni-Mol & \textbf{85.7 $\pm$ 0.2} & 72.9 $\pm$ 0.6 & \textbf{80.8 $\pm$ 0.3} \\
\midrule
$\texttt{agg}+\texttt{I}$ (ours) & 82.39 $\pm$ 0.53 & \textbf{74.08 $\pm$ 0.0} & 76.95 $\pm$ 0.34 \\
\bottomrule
\end{tabular}
\end{table*}         \begin{table*}[t]
\centering
\small
\caption{\textbf{LRGB peptides.} Test average precision (multi-label classification, mean $\pm$ std) and test mean absolute error (multi-task regression, mean $\pm$ std). Numbers are taken from the cited papers. We include every model that reports a value on at least one of the three datasets. Bold entries mark the single best reported value in a column.}
\label{tab:bm_peptides}
\setlength{\tabcolsep}{4pt}
\begin{tabular}{l c c}
\toprule
 & \textbf{Peptides-func} & \textbf{Peptides-struct} \\
Model & {\itshape AP $\uparrow$} & {\itshape MAE $\downarrow$} \\
\midrule
\multicolumn{3}{l}{\citet{dwivedi22a}} \\
\addlinespace[2pt]
GCN & 0.5930 $\pm$ 0.0023 & 0.3496 $\pm$ 0.0013 \\
GCNII & 0.5543 $\pm$ 0.0078 & 0.3471 $\pm$ 0.0010 \\
GINE & 0.5498 $\pm$ 0.0079 & 0.3547 $\pm$ 0.0045 \\
GatedGCN & 0.5864 $\pm$ 0.0077 & 0.3420 $\pm$ 0.0013 \\
GatedGCN+RWSE & 0.6069 $\pm$ 0.0035 & 0.3357 $\pm$ 0.0006 \\
Transformer+LapPE & 0.6326 $\pm$ 0.0126 & 0.2529 $\pm$ 0.0016 \\
SAN+LapPE & 0.6384 $\pm$ 0.0121 & 0.2683 $\pm$ 0.0043 \\
SAN+RWSE & 0.6439 $\pm$ 0.0075 & 0.2545 $\pm$ 0.0012 \\
\midrule
\multicolumn{3}{l}{\citet{rampasek22a}} \\
\addlinespace[2pt]
GPS & 0.6535 $\pm$ 0.0041 & 0.2500 $\pm$ 0.0005 \\
\midrule
\multicolumn{3}{l}{\citet{toenshoff2024where}} \\
\addlinespace[2pt]
GCN (retuned) & 0.6860 $\pm$ 0.0050 & 0.2460 $\pm$ 0.0007 \\
GINE (retuned) & 0.6621 $\pm$ 0.0067 & 0.2473 $\pm$ 0.0017 \\
GatedGCN (retuned) & 0.6765 $\pm$ 0.0047 & 0.2477 $\pm$ 0.0009 \\
GPS (retuned) & 0.6534 $\pm$ 0.0091 & 0.2509 $\pm$ 0.0014 \\
CRaWl & 0.7074 $\pm$ 0.0032 & 0.2506 $\pm$ 0.0022 \\
DRew & \textbf{0.7150 $\pm$ 0.0044} & 0.2536 $\pm$ 0.0015 \\
Exphormer & 0.6527 $\pm$ 0.0043 & 0.2481 $\pm$ 0.0007 \\
GRIT & 0.6988 $\pm$ 0.0082 & 0.2460 $\pm$ 0.0012 \\
Graph ViT & 0.6942 $\pm$ 0.0075 & \textbf{0.2449 $\pm$ 0.0016} \\
G-MLPMixer & 0.6921 $\pm$ 0.0054 & 0.2475 $\pm$ 0.0015 \\
\midrule
$\texttt{agg}+\texttt{I}$ (ours) & 0.6874 $\pm$ 0.0022 & 0.2476 $\pm$ 0.0002 \\
\bottomrule
\end{tabular}
\end{table*}         \begin{table*}[t]
\centering
\small
\caption{\textbf{Flow datasets.} Test mean absolute error (mean or mean $\pm$ std). Numbers are taken from the cited paper. We include every model that reports a value on at least one of the three datasets. Bold entries mark the single best reported value in a column.}
\label{tab:bm_flow}
\setlength{\tabcolsep}{4pt}
\begin{tabular}{l c c c}
\toprule
 & \multicolumn{3}{c}{\textit{MAE $\downarrow$}}\\
\cmidrule(lr){2-4}
Model & \textbf{Flow\_easy} & \textbf{Flow\_medium} & \textbf{Flow\_hard} \\
\midrule
\multicolumn{4}{l}{\citet{stoll26a}} \\
\addlinespace[2pt]
GIN & 3.44 & 9.60 & 9.51 \\
Graph transformer & 4.27 & 6.38 & 6.48 \\
\midrule
$\texttt{agg}+\texttt{I}$ (ours) & \textbf{1.83 $\pm$ 0.01} & \textbf{5.30 $\pm$ 0.12} & \textbf{5.26 $\pm$ 0.03} \\
\bottomrule
\end{tabular}
\end{table*}         \clearpage
    
\begin{figure}
    \centering
    \includegraphics[width=0.90\linewidth]{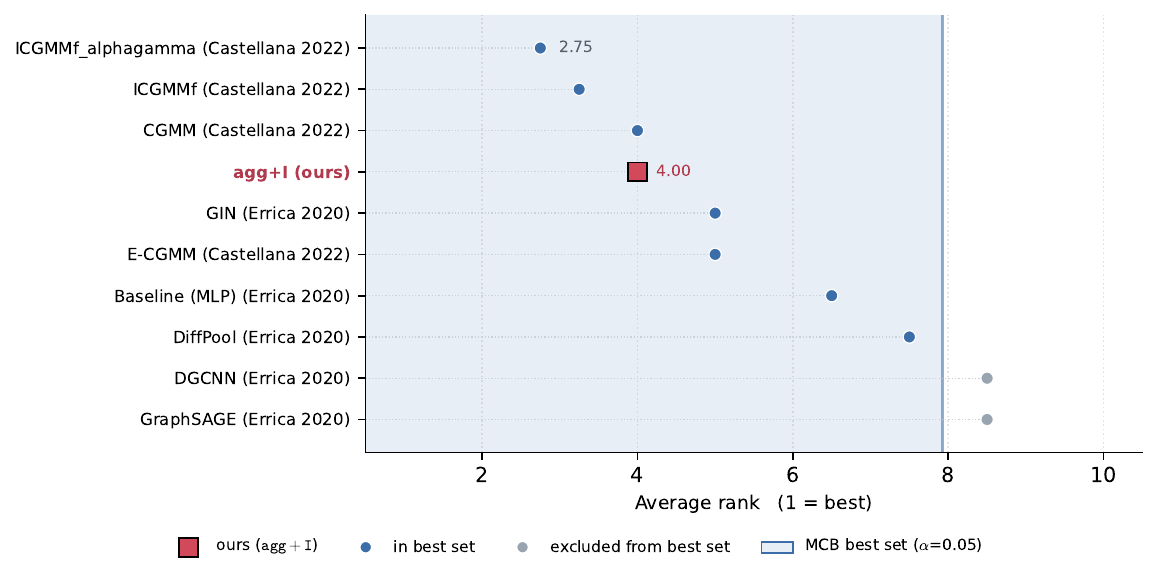}
    \caption{Multiple comparisons with the best (MCB) for the TU datasets of \Cref{tab:bm_tu}. Points show the average rank of each method over the $n=4$ datasets for the $k=10$ methods that report all four datasets. The shaded band is the MCB best set at $\alpha=0.05$ (critical distance $5.17$): methods inside it (blue) cannot be ruled out as the best, methods outside it (grey) are significantly worse than the best. Our $\texttt{agg}+\texttt{I}$ is shown in red.}
    \label{fig:mcb_tu}
\end{figure}

\begin{figure}
    \centering
    \includegraphics[width=0.90\linewidth]{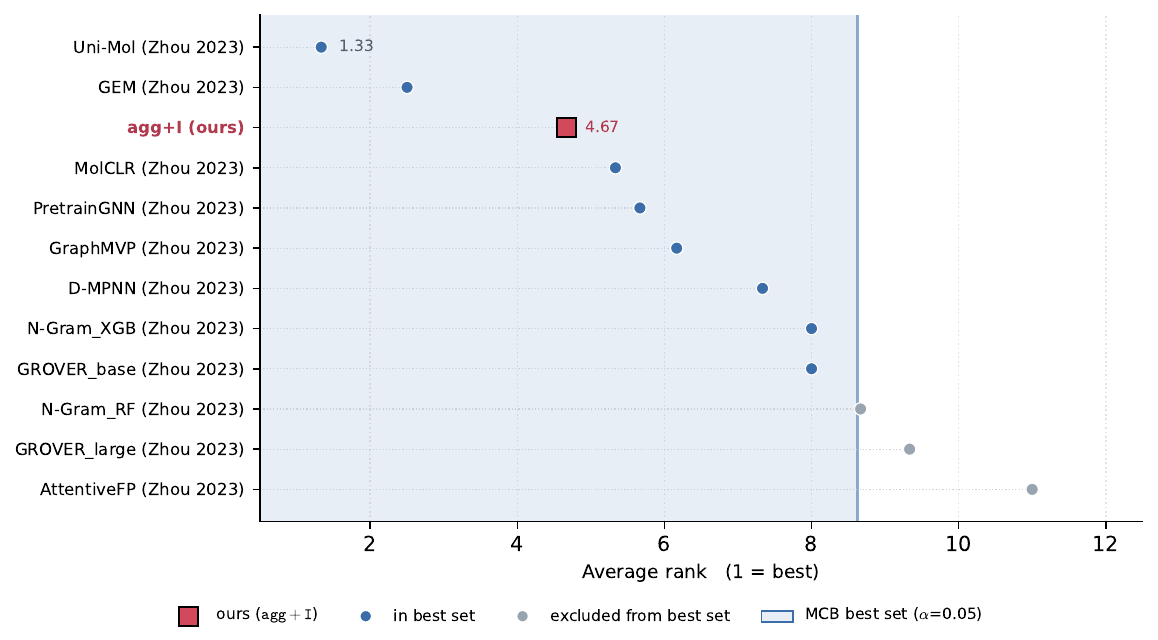}
    \caption{Multiple comparisons with the best (MCB) for the MoleculeNet datasets of \Cref{tab:bm_molecules}. Points show the average rank of each method over the $n=3$ datasets for all $k=12$ methods. The shaded band is the MCB best set at $\alpha=0.05$ (critical distance $7.29$): methods inside it (blue) cannot be ruled out as the best, methods outside it (grey) are significantly worse than the best. Our $\texttt{agg}+\texttt{I}$ is shown in red.}
    \label{fig:mcb_molecules}
\end{figure}

\begin{figure}
    \centering
    \includegraphics[width=0.90\linewidth]{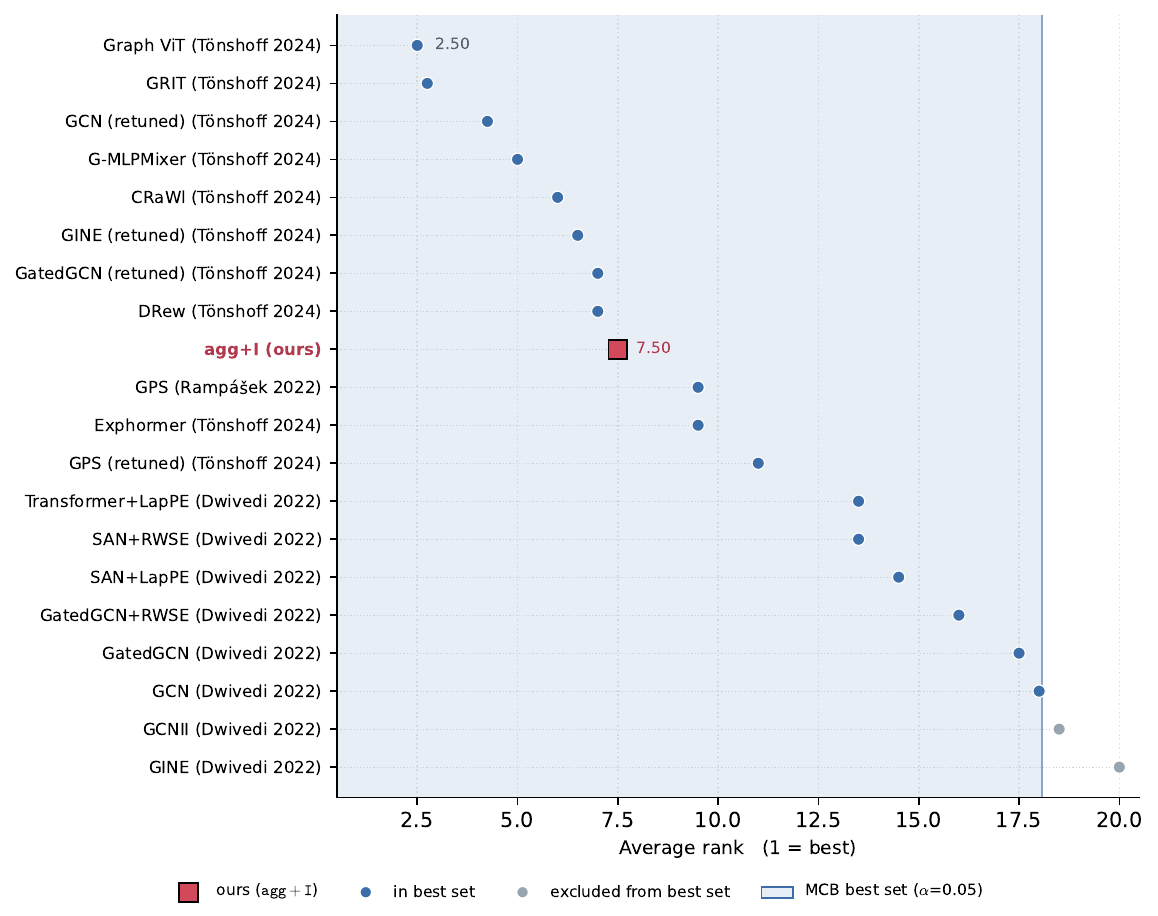}
    \caption{Multiple comparisons with the best (MCB) for the LRGB datasets of \Cref{tab:bm_peptides}. Points show the average rank of each method over the $n=2$ datasets for all $k=20$ methods. The shaded band is the MCB best set at $\alpha=0.05$ (critical distance $15.56$): methods inside it (blue) cannot be ruled out as the best, methods outside it (grey) are significantly worse than the best. Our $\texttt{agg}+\texttt{I}$ is shown in red. With only two datasets, this comparison is descriptive only.}
    \label{fig:mcb_peptides}
\end{figure}

    \paragraph{Multiple comparisons with the best.}
        \label{app:mcb}
        We compare methods within each benchmark table using Hsu's multiple
        comparisons with the best~\citep[MCB]{hsu84a, Horrace20a} on average
        ranks~\citep{demsar06a}; cf.\ \Cref{fig:mcb_tu,fig:mcb_molecules,fig:mcb_peptides}.
On every dataset, methods are ranked by their reported mean, and methods with missing entries are excluded from that table's analysis.
A method belongs to the \emph{best set} if its average rank is within the critical distance~$c$ from the best average rank.
Following \citet{hsu84a} and \citet{demsar06a}, the critical distance is $c = d_{\alpha}\sqrt{k(k+1)/(6n)}$, where $k$ is the number of methods, $n$ the number of datasets, and $d_{\alpha}$ the one-sided Dunnett constant for $k-1$ comparisons at $\alpha=0.05$.
With only $n=2$--$4$ datasets per table, $c$ is large and MCB has limited power: Membership in the best set means that a method is not significantly worse than the best.
 
\clearpage
\section{Invariants reveal obstructions to multi-task learning}
    \label{sec:Multi-Task Performance}

    We investigate whether the structural differences between datasets observed in \Cref{sec:Heterogeneity} can \emph{impact} multi-task learning performance.
Unlike most multi-task and pretraining approaches on graphs~\citep{qiu20a, frasca25a, hu20b, liu24a, du20a, yu20a}, we avoid a separate pretraining objective and train directly on two downstream tasks simultaneously, so that structural differences directly act on the prediction tasks.
To our knowledge, neither multi-task training across graph domains nor gradient alignment during multi-task training on graphs has been studied before.
We address this gap by assessing the correlation between structural differences and gradient alignment, which is known to be an important indicator of the learning process~\citep{du20a, yu20a}, as well as structural differences and the performance outcomes.
We measure structural difference as the accuracy of pairwise meta-classification: an accuracy near $50\%$ indicates structurally similar datasets, a high accuracy structurally dissimilar ones.

    \paragraph{Model and setup.}
        We use a shared Graph Isomorphism Network~\citep[GIN]{xu18a} graph-encoder, a message-passing network~\citep{Corso24a}, followed by sum pooling and one task head per dataset~(hyperparameters in \cref{tab:gin_hyperparameters}).
We train on all $66$ pairs of $12$ datasets~(BACE, BBBP, HIV, Lipo, ZINC, QM9, NCI1, NCI109, PROTEINS, Peptides-func, Peptides-struct, and CO\_BA\_small), which span structurally similar and dissimilar pairs, i.e., both frequently confused and easily distinguished datasets in the meta-classification confusion matrix~(\cref{fig:confusion_all}).
In each step, one batch per dataset is passed through the shared encoder, and gradients are computed separately for both batches to measure their alignment in the encoder.
All models are trained from scratch for each pair, repeated for $5$ random seeds.
To isolate structure and allow training across domains, node features are replaced by constant unit features.
We train from scratch for each pair and each of $5$ seeds; an epoch iterates over the larger dataset, while cycling through the smaller one, and training stops once the validation loss on both datasets has plateaued.
We evaluate on each test set at the epoch with minimal validation loss.

\begin{figure}[h]
    \centering
    \includegraphics[width=1.0\textwidth]{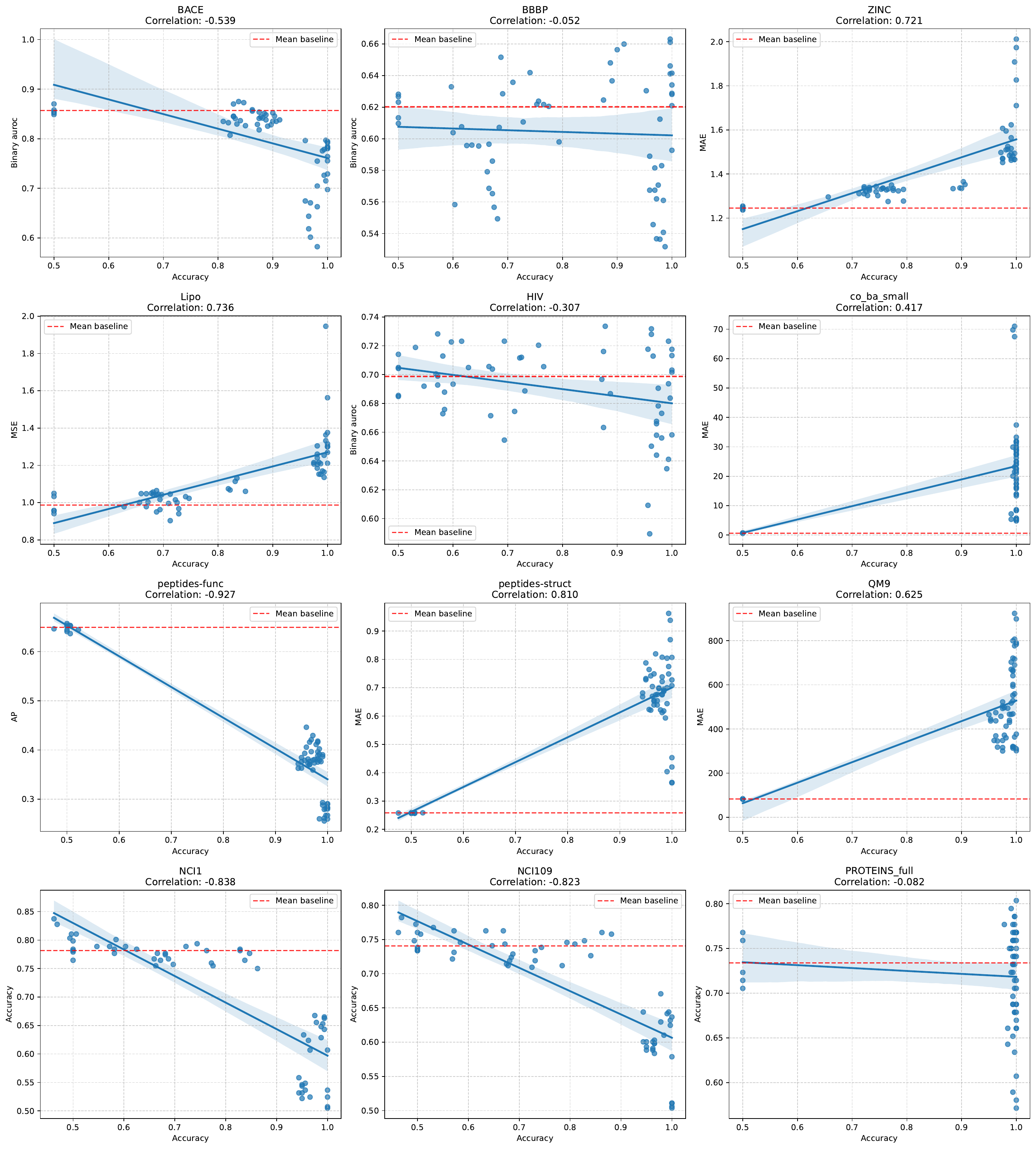}
    \caption{\textbf{Multi-task test performance, all datasets.} Correlation between meta-accuracy and test performance after multi-task learning. Higher accuracy is correlated with lower performance after multi-task learning on all datasets except BBBP and PROTEINS, which show no clear correlation (notice that for some metrics, lower values are preferable). We repeat multi-task training and classification for $5$ seeds for each pair of datasets, showing one seed as one dot.
    }
    \label{fig:grid_meta_acc_vs_test}
\end{figure}

\begin{figure}[tbp]
    \centering
    \includegraphics[width=1.0\textwidth]{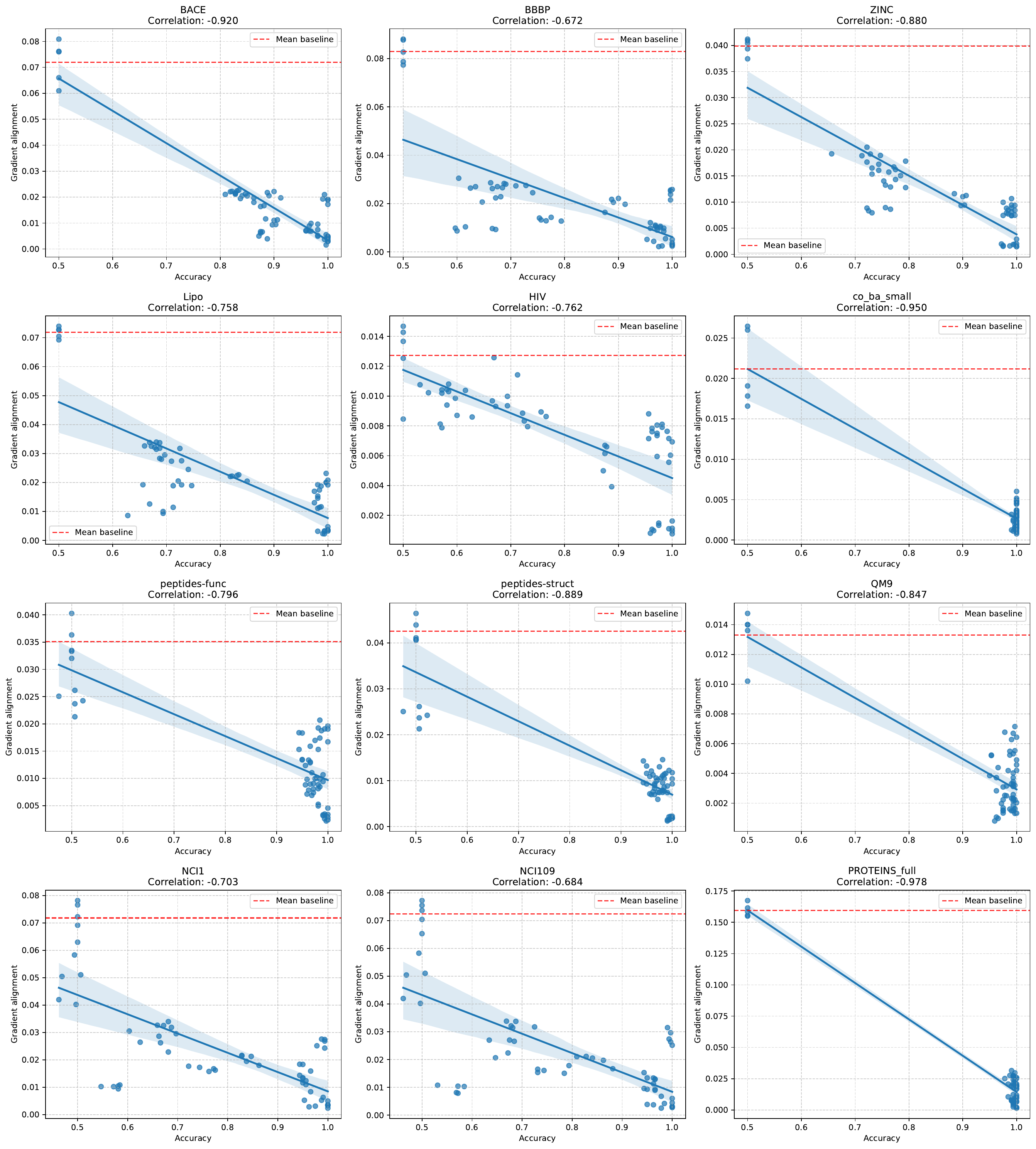}
    \caption{\textbf{Multi-task gradient alignment, all datasets.} Correlation between meta-accuracy and gradient alignment during multi-task learning. Higher accuracy is consistently correlated with lower gradient alignment. We repeat multi-task training and classification for $5$ seeds for each pair of datasets, showing one seed as one dot.
    }
    \label{fig:grid_meta_acc_vs_alignment}
\end{figure}

    \paragraph{Results.}
        Higher meta-classification accuracy coincides with worse multi-task test performance on $10$ of $12$ datasets, with no clear correlation for BBBP and PROTEINS~(\Cref{fig:grid_meta_acc_vs_test}), and with lower gradient alignment on all $12$ datasets~(\Cref{fig:grid_meta_acc_vs_alignment}): the more structurally distinguishable two datasets are, the worse they tend to train together.
Although we do not observe strictly negative alignment~\cite{du20a, hu20b}, we hypothesize that the reduced alignment impedes multi-task performance.
Invariants can thus serve as a diagnostic for obstructions to joint training or pretraining.
As a limitation, each pair introduces both a second graph distribution and a second task, and we do not disentangle the two effects.

                  
\end{document}